\documentclass[11pt]{article}

\usepackage[preprint]{acl}

\usepackage{times}
\usepackage{latexsym}

\usepackage[T1]{fontenc}

\usepackage[utf8]{inputenc}

\usepackage{microtype}

\usepackage{inconsolata}

\usepackage{graphicx}
\usepackage{tabularx}
\usepackage{amsmath}
\usepackage{amssymb}
\definecolor{myblue}{HTML}{2973B2}
\definecolor{mypurple}{HTML}{c0165f}

\usepackage{multirow}
\usepackage{array}

\usepackage{xspace}
\newcommand{\method}{Sherry\xspace}

\title{\raisebox{-0.1cm}{\includegraphics[height=1.2em]{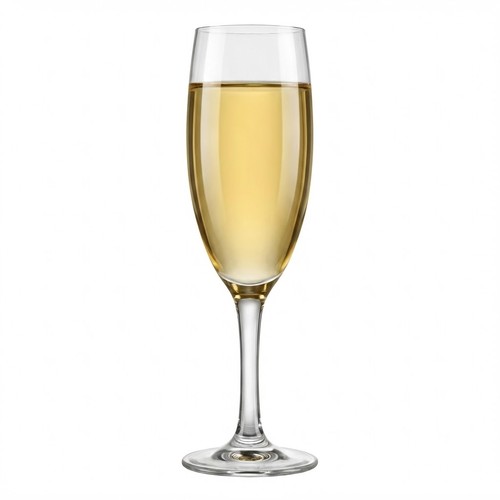}} \method: Hardware-Efficient 1.25-Bit Ternary Quantization via Fine-grained Sparsification}

\author{\textbf{Hong Huang}$^1$\thanks{Work with Tencent}\ \
        \textbf{Decheng Wu}$^2$\ \ 
        \textbf{Qiangqiang Hu}$^2$\ \
        \textbf{Guanghua Yu}$^2$\ \ 
        \textbf{Jinhai Yang}$^1$\\ 
        \textbf{Jianchen Zhu}$^2$\ \ 
        \textbf{Xue Liu}$^{3}$\ \ 
        \textbf{Dapeng Wu}$^1$\\
  $^1$City University of Hong Kong
  \quad
  $^2$Tencent
  \quad 
  $^3$McGill University \\
}

\begin{document}
\maketitle

\begin{abstract}
The deployment of Large Language Models (LLMs) on resource-constrained edge devices is increasingly hindered by prohibitive memory and computational requirements. While ternary quantization offers a compelling solution by reducing weights to $\{-1, 0, +1\}$, current implementations suffer from a fundamental misalignment with commodity hardware. Most existing methods must choose between 2-bit aligned packing, which incurs significant bit wastage, or 1.67-bit irregular packing, which degrades inference speed. 
To resolve this tension, we propose \textbf{\method}, a hardware-efficient ternary quantization framework. 
\method introduces a 3:4 fine-grained sparsity that achieves a regularized 1.25-bit width by packing blocks of four weights into five bits, restoring power-of-two alignment. Furthermore, we identify \textit{\textbf{weight trapping}} issue in sparse ternary training, which leads to representational collapse. To address this, Sherry introduces \textbf{Arenas}, an annealing residual synapse mechanism that maintains representational diversity during training. Empirical evaluations on LLaMA-3.2 across five benchmarks demonstrate that \method matches state-of-the-art ternary performance while significantly reducing model size. Notably, on an Intel i7-14700HX CPU, our 1B model achieves zero accuracy loss compared to SOTA baselines while providing 25\% bit savings and 10\% speed up. The code is available at \url{https://github.com/Tencent/AngelSlim}.
\end{abstract}

\section{Introduction}
Large language models (LLMs) have demonstrated remarkable capabilities across massive applications~\cite{wu2023brief, floridi2020gpt}. However, the increasing concerns regarding data privacy, the necessity for offline functionality, the prohibitive latency, and the cost of cloud-based inference~\citep{yao2024survey, liagkou2024cost} have driven a pressing need to deploy LLMs on resource-constrained edge devices.
\begin{figure}[t]
    \centering    
    \includegraphics[width=0.9\linewidth]{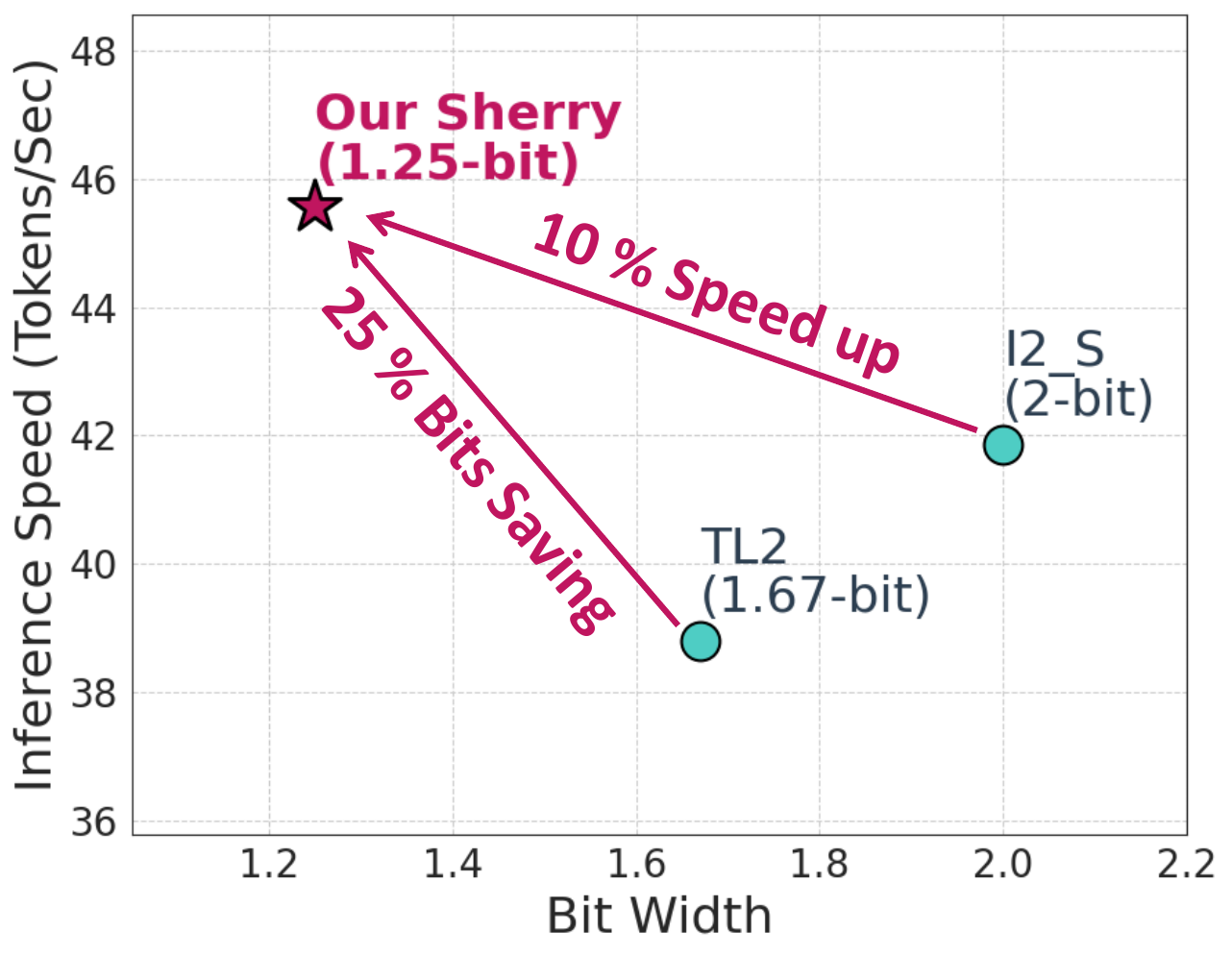}
    \vspace{-1em}
    \caption{Comparison of different packing strategies for ternary quantization in efficiency.}
    \label{fig: first}
    \vspace{-1.5em}
\end{figure}

Weight quantization~\citep{hubara2018quantized,liu2023qllm} is a practical technique for enabling on-device deployment by reducing model footprints and computation through lower-precision numerical representations. Nevertheless, most existing quantization methods~\citep{dettmers2024qlora, lin2023awq, frantar2022gptq} are optimized for server-grade GPUs that support complex hardware primitives, such as mixed-precision multiplication. Thus, these methods are often unsuitable for heterogeneous edge and mobile hardware, which prioritize simplified, widely supported operations that avoid specialized hardware dependencies.

\begin{figure*}[t]
    \centering
    \includegraphics[width=0.95\linewidth]{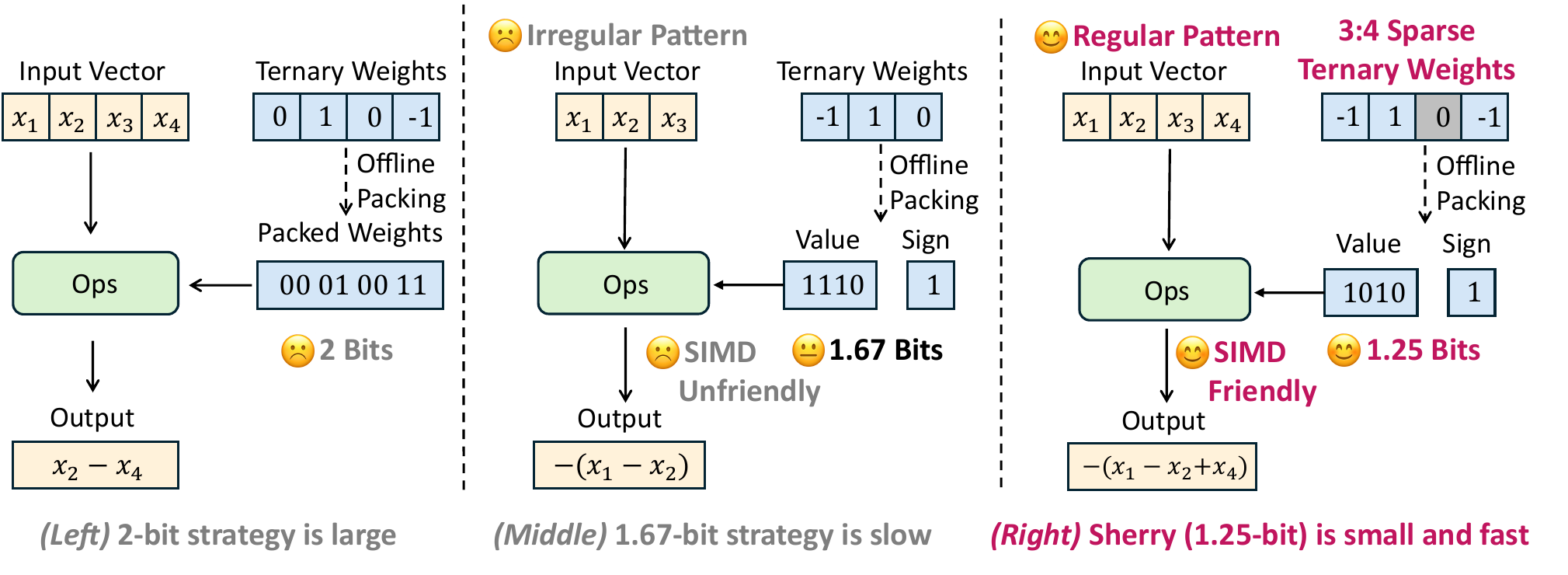}
    \vspace{-1em}
   \caption{ %
    \textit{(Left)} \textbf{2-bit strategy} packs each weight into 2 bits to maintain alignment, resulting in large bit wastage.
    \textit{(Middle)} \textbf{ 1.67-bit strategy} packs 3 weights into 5 bits, introducing SIMD-unfriendly 3-way patterns, leading to slow speed. 
    \textit{(Right)} \textbf{Our Sherry} enforces a 3:4 sparsity and packs 4 weights into 5 bits,  introducing SIMD-friendly 4-way patterns, achieving a small 1.25-bit width and faster inference speed. More details are shown in Fig~\ref{fig: sys}.}
    \label{fig: main}
    \vspace{-1.5em}
\end{figure*}

Ternary quantization~\citep{li2016ternary, liu2023ternary} provides a compelling paradigm for edge-based LLM inference by constraining weights to the set $\{-1, 0, +1\}$. Supported by the Lookup Table (LUT)-based engines~\cite{huang2025tenet, wei2025t, wang2025bitnetcpp, nie2025elutq}, ternary quantization transforms costly floating-point multiplications into hardware-efficient additions, as shown in Fig~\ref{fig: sys}. This intrinsic compatibility makes ternary quantization a promising solution to bridge the gap between high-performance LLMs and the strict resource constraints of edge devices.

However, existing ternary quantization methods~\citep{ ma2025bitnet, wang2025bitnet, huang2025tequila} suffer from practical inefficiencies due to the misalignment between non-standard ternary bit-widths and standard hardware architectures. Current implementations generally resort to two suboptimal strategies for LUT-based execution: (1) \textbf{2-bit strategy} that packs each weight into 2 bits~\citep{wei2025t}, as shown in Fig.~\ref{fig: main} \textit{(left)}, offering no memory savings over standard INT2 quantization; or (2) \textbf{1.67-bit strategy} that packs three weights into five bits~\citep{wang2025bitnetcpp}, as shown in Fig.~\ref{fig: main} \textit{(middle)}. While the 1.67-bit strategy offers improved density, its 3-way grouping is fundamentally incompatible with the power-of-two vector lanes of modern Single Instruction Multiple Data (SIMD) units, frequently leading to a slower inference speed compared to the 2-bit strategy. Consequently, \textbf{existing ternary methods are forced to trade off between bit width and inference speed}, preventing them from achieving the full theoretical benefits of ternary weights.

To address these limitations, we propose \textbf{Sherry}, a novel \underline{\textbf{S}}parse \underline{\textbf{h}}ardware-\underline{\textbf{e}}fficient t\underline{\textbf{er}}na\underline{\textbf{r}}y quantization framework that achieves 1.25-bit width while maintaining superior inference speed. Our key insight is that the inherent sparsity of ternary models can be strategically structured to reconcile the tension between storage density and computational regularity. Specifically, \method enforces an optimal 3:4 fine-grained sparsity constraint: within every contiguous block of four weights, exactly three are quantized to non-zero values ($\pm 1$), and one is fixed to zero. This structured constraint enables an optimal packing strategy where each 4-weight block is stored in a compact 5-bit representation. Crucially, this block-based approach restores the power-of-two alignment required by modern SIMD units, allowing for parallel processing in regularized hardware operations.

Furthermore, directly integrating 3:4 sparsity into ternary training often results in performance degradation. We attribute this to \textit{\textbf{weight trapping}: weights accumulate in localized regions due to \textbf{gradient homogenization}, leading to representational collapse,} as shown in Fig.~\ref{fig: problem} and~\ref{fig: more_distribution}. To address this, \method introduces \textbf{Arenas}\footnote{"Arenas" is also the name of the sandy soil used to grow grapes for sweet Sherry wine.}, an \underline{\textbf{A}}nnealing \underline{\textbf{re}}sidual sy\underline{\textbf{na}}p\underline{\textbf{s}}e module. By injecting heterogeneous gradients during the training phase, this module breaks the gradient homogenization and maintains an expressive and diverse weight distribution for \method. %

We evaluate the efficacy and efficiency of \method across five standard benchmarks using the LLaMA-3.2~\citep{touvron2023llama} model family. Our empirical results demonstrate that \method matches performance with state-of-the-art (SOTA) ternary quantization baselines while necessitating a substantially lower bit width. Notably, on an Intel i7-14700HX CPU, \method achieves a 16\% reduction in model size and a $10\%$ inference speed up compared to SOTA 1B ternary LLM, with zero accuracy degradation. These findings validate that \method provides a promising hardware-efficient solution for deploying LLMs on resource-constrained platforms.

\section{Background and Challenge}

\subsection{Ternary Quantization}
Ternary quantization is an extreme weight compression paradigm that constrains model parameters to the discrete set $\{-1, 0, +1\}$. Taking per-channel quantization as an example, for a full-precision weight matrix $W \in \mathbb{R}^{d_{in} \times d_{out}}$, where  $d_{in}$ is the number of input channels and $d_{out}$ is the number of output channels, the general ternary quantization function $Q(\cdot)$ is defined as:
\begin{equation}
    Q(W) = T\alpha, \quad T_{i,j} = \begin{cases} 
    +1, & \text{if } W_{i,j} > \Delta_j; \\ 
    0, & \text{if } |W_{i,j}| \le \Delta_j; \\ 
    -1, & \text{if } W_{i,j} < -\Delta_j, 
    \end{cases} 
    \label{eq:tquant}
\end{equation}
where $T \in \{-1, 0, +1\}^{d_{in} \times d_{out}}$ is the ternary weight matrix, $\alpha \in \mathbb{R}^{d_{out}}$ represents the scaling factors\footnote{In this paper, we denote the multiplication between matrix $T$ and vector $\alpha$ as element-wise, \textit{i.e.}, $[T\alpha]_{i,j} = T_{i,j}\alpha_j$.}, and $\Delta \in \mathbb{R}^{d_{out}}$ denotes the quantization thresholds. A substantial body of research has investigated the optimal determination of $\alpha$ and $\Delta$, which are discussed in detail in Sec.~\ref{sec: baseline}.

To eliminate weight-activation multiplications during inference, \textbf{Lookup Table (LUT)-based engines}~\cite{wang2025bitnetcpp, wei2025t} have been developed, demonstrating superior efficiency over traditional multiplication-based engines. As illustrated in Fig.~\ref{fig: sys}, the engine segments the inputs and pre-computes a localized lookup table. For each segment, the corresponding ternary weight index is used to retrieve pre-computed results from the table. This paradigm provides a promising pathway for ternary LLMs by fully replacing complex floating-point multiplications with highly efficient additions and memory lookups.

\subsection{Quantization-Aware Training (QAT)}
Due to the aggressive nature of ternary compression, QAT is essential to recover model fidelity. In the forward pass, full-precision weights $W$ are dynamically quantized via $Q(\cdot)$ in Eq.~\ref{eq:tquant}. During the backward pass, since $Q(\cdot)$ is non-differentiable, we employ the Straight-Through Estimator (STE)~\citep{zhu2016trained, chen2024ternaryllm} to approximate gradients. For an input matrix $X \in \mathbb{R}^{d_t\times d_{in}}$, where $d_t$ is the number of tokens, the forward pass and gradients for loss $L$ are:
\begin{equation}
Y = XT\alpha, \quad   \frac{\partial L}{\partial W} \approx X^\top\frac{\partial L}{\partial Y}.
\label{eq:ste}
\end{equation}
After training, the weights $W$ are discarded, leaving the ternary weights $T$ and scaling factors $\alpha$.

\subsection{Challenge}
Although ternary quantization has a theoretical lower bound of $1.58$ bits ($\log_2 3$), contemporary implementations encounter significant architectural friction when deployed on commodity hardware. Most current methods resort to two suboptimal strategies: \textbf{(1)  2-bit Strategy:} This approach pads each ternary weight into 2 bits~\citep{wei2025t}, as shown in Fig.~\ref{fig: main} \textit{(left)}. While it preserves computational regularity and aligns with SIMD vector lanes, it effectively nullifies the storage advantages of the ternary set compared to 2-bit integer quantization.
\textbf{(2) 1.67-bit Strategy:} This scheme packs three weights into 5-bit blocks~\citep{wang2025bitnetcpp}, as shown in Fig.~\ref{fig: main} \textit{(middle)}. However, this introduces severe arithmetic inefficiencies because modern hardware accelerators are optimized for $2^n$ operand groupings. In practice, this misalignment often results in slower inference speeds than the 2-bit strategy.

Alternative strategies, such as packing 5 weights into 8 bits ($1.6$-bit), exponentially increase the size of the lookup tables, rendering them unsuitable for LUT-based edge engines. Consequently, existing ternary quantization remains trapped in a trade-off between memory footprint and inference speed, preventing it from realizing its full theoretical benefits in practical edge deployments.

\begin{figure*}[t]
    \centering    \includegraphics[width=0.95\linewidth]{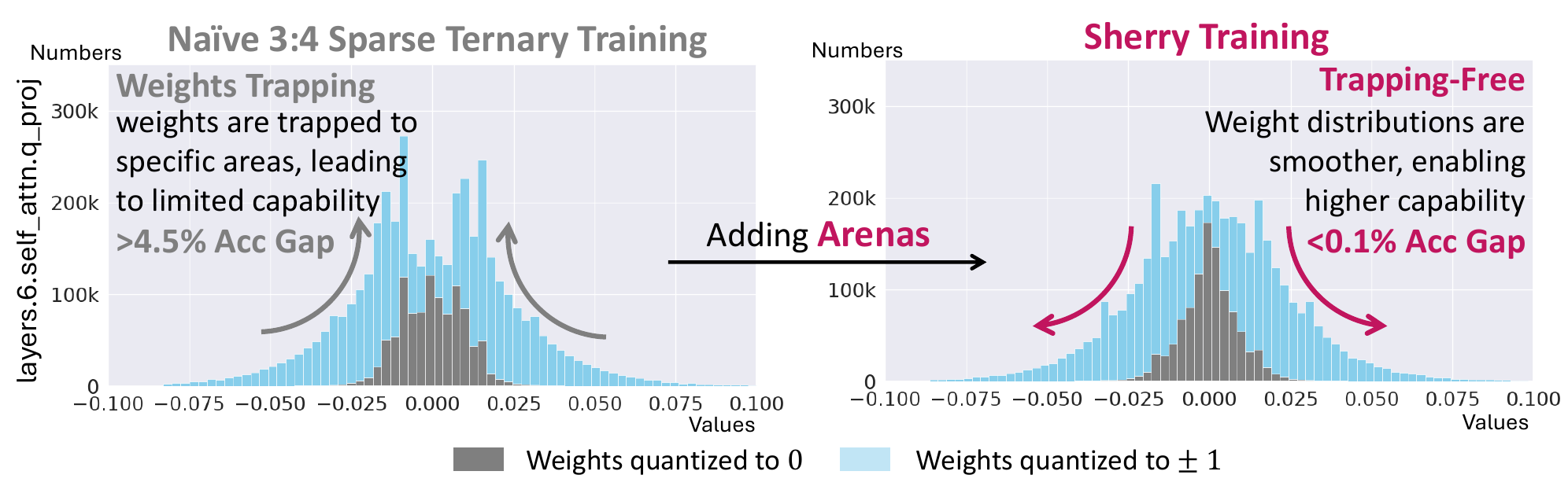}
    \vspace{-1em}
    \caption{Comparison of weight distributions for LLaMA-1B under different quantization schemes. 
\textit{(Left)} Naive 3:4 sparse ternary training exhibits \textbf{weight trapping}, where values collapse into a binary-like polarization, leading to a significant accuracy gap compared to dense models. 
\textit{(Right)} Our \method utilizes the Arenas module to achieve a trap-free distribution,  bridging the performance gap to dense ternary models.}
    \label{fig: problem}
    \vspace{-1.5em}
\end{figure*}

\section{Proposed \method}
\subsection{3:4 Sparse Ternary Quantization}
The primary objective of \method is to achieve substantial bit savings without compromising inference throughput. Building on the observation that ternary models exhibit inherent sparsity~\cite{liu2023ternary}, we formalize this characteristic by adopting an N:M structured sparsity constraint. This enforces a rigid pruning pattern where exactly $N$ non-zero elements are permitted within every contiguous block of $M$ weights.

While traditional $N:M$ sparsity (e.g., the 2:4 pattern) is typically coupled with specific GPU-vendor kernels~\citep{zhou2021learning}, \method leverages a multiplication-free LUT engine to decouple the sparsity constraint from specialized hardware primitives. This architectural freedom allows us to explore a 3:4 structured sparsity pattern, which achieves a highly efficient 1.25-bit width. 
The optimization goal of \method is to minimize the $L_2$ reconstruction error between the full-precision weights $W$ and the sparse ternary representation $T\alpha$, subject to the 3:4 sparsity constraint. Let $W_{b:b+3, j} = (W_{b,j}, W_{b+1,j},W_{b+2,j},W_{b+3,j})$ denote a contiguous vector. The objective is formulated as\footnote{In this paper, we denote $W_{:, j} \in \mathbb{R}^{d_{in}}$ as the $j$-th column o matrix  $W\in \mathbb{R}^{d_{in}\times d_{out}}$, \textit{i.e.,} $W_{:, j} = (W_{1,j},\dots W_{d_{in},j})$}:
\begin{equation}
\begin{aligned}
\min_{T, \alpha} & \sum_{j=1}^{d_{out}} \|W_{:, j} - T_{:, j} \alpha_j \|_2^2 \\
\text{s.t. } & T_{i,j} \in \{-1, 0, +1\}, \\
& \forall b\in \{1, 5,\dots, d_{in}-4 \}: \|T_{b:b+3, j}\|_0 = 3,
\end{aligned}
\label{eq: sparse_objective}
\end{equation}
where $\|\cdot\|_0 = 3$ enforces the 3:4 sparsity by ensuring exactly three non-zero values per block. 

The optimal solution to the objective in Eq.~\ref{eq: sparse_objective} is obtained via a greedy Sparse-AbsMean strategy. For each block $W_{b:b+3, j}$, we prune the element with the smallest absolute magnitude and assign ternary values to the remaining three. Formally, for each block $i\in[b,b+3]$, the optimal ternary element $T_{i,j}$ is given by:
\begin{equation}
T^*_{i,j} = \begin{cases} 
0, & \text{if } i = \arg\min |W_{i,j}|; \\ 
\text{sign}(W_{i,j}), & \text{otherwise}.
\end{cases}
\label{eq:optimal_T}
\end{equation}
Given this optimal ternary matrix $T$, the optimal scaling factor $\alpha_j$ for the $j$-th output channel is calculated as the mean absolute value of the non-pruned weights:
\begin{equation}
\alpha^*_j = \frac{4}{3 d_{in}} \sum_{i \in \mathcal{S}_j} |W_{i,j}|,
\label{eq:optimal_alpha}
\end{equation}
where $\mathcal{S}_j = \{i \mid T_{i,j} \neq 0\}$ denotes the set of active (non-zero) indices in the $j$-th column. The proof of the optimal is presented in Sec.~\ref{sec: proof}.

Although rarely used in floating-point contexts, 3:4 sparsity represents \textbf{an ideal "sweet spot"} for ternary quantization across four dimensions:\\
\begin{figure*}[!t]
\centering
\begin{minipage}[t]{0.48\textwidth}
\centering
\includegraphics[width=0.9\linewidth]{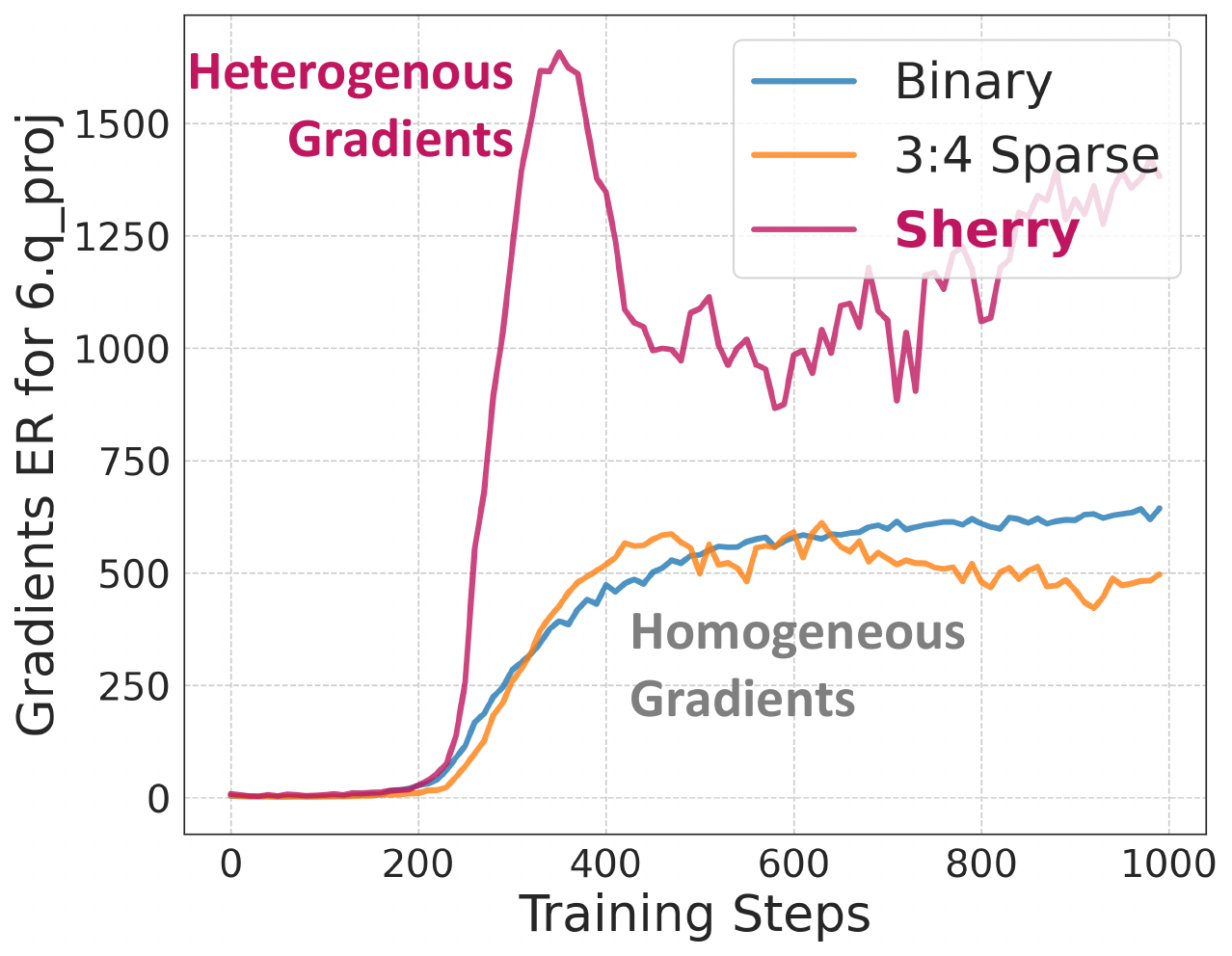}
    \vspace{-1em}
    \caption{The effective ranks (ER) of the gradients during training. Both Binary and 3:4 sparse ternary training have relatively low ER due to gradient homogenization.}
    \label{fig: er}
    \vspace{-1em}
\end{minipage}
\hspace{0.2em}
\begin{minipage}[t]{0.48\textwidth}
\centering
\includegraphics[width=0.9\linewidth]{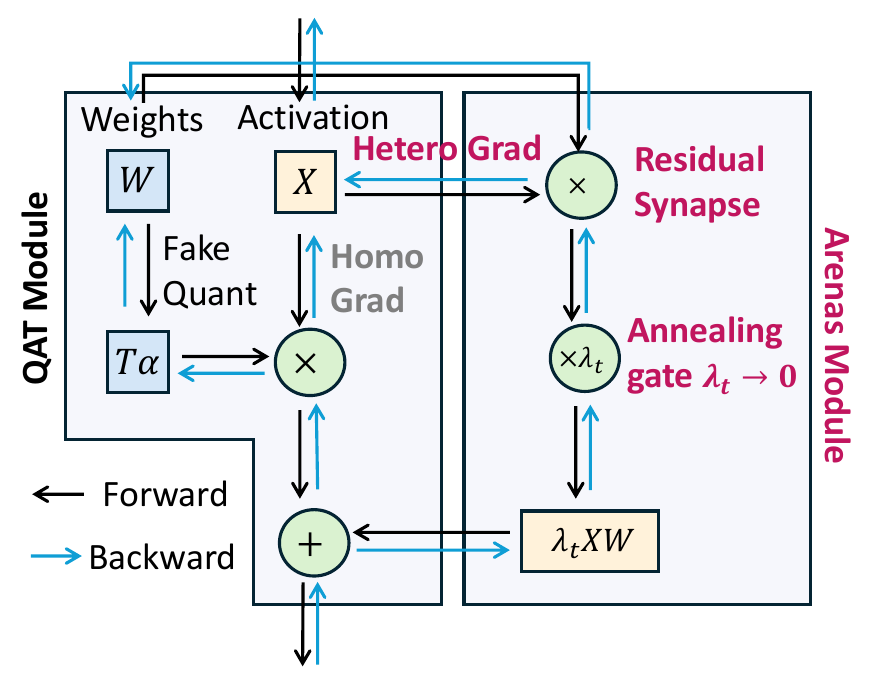}
    \vspace{-1em}
    \caption{The overview of the Arenas module with QAT. The Arenas module injects the heterogeneous gradients through a residual synapse with an annealing gate.}
    \label{fig: arenas}
    \vspace{-1em}
\end{minipage}
\end{figure*}
\textbf{(1) SIMD-friendly Alignment:} The choice of $M=4$ ensures power-of-two alignment with activation segments, which is critical for SIMD vector lane loading. This eliminates the complex bit-shuffling overhead characteristic of 1.67-bit (3-way) packing schemes.\\
\textbf{(2) Safe Sparsity Margin:} Prior research~\cite{zhu2016trained} indicates that ternary quantization performance undergoes severe degradation when the sparsity ratio exceeds 50\%. 3:4 structured scheme maintains a 25\% sparsity level within the safe margin required to preserve model expressive capacity.\\
\textbf{(3) Optimal Bit-State Utilization:} For $M=4$, enforcing $N=3$ non-zero elements yields exactly $\binom{4}{3} \times 2^3 = 32$ unique permutations. This mathematically saturates a 5-bit index ($2^5 = 32$), ensuring maximum bit-state utilization without bit wastage.\\
\textbf{(4) Compatibility with LUT-Inference:} The 3:4 pattern is natively compatible with the LUT-based ternary inference engine, as shown in Fig~\ref{fig: sys}. Given the 128-bit constraints of standard SIMD register instructions (e.g., AVX2 vpshufb), the maximum capacity for a single-instruction lookup table is 16 bytes. By utilizing mirror-symmetry of ternary states~\cite{ma2025bitnet, wei2025t}, \method splits the 5-bit representation into 1 sign bit and 4 index bits, perfectly fitting within search limits.

In summary, the 3:4 format represents an intersection of hardware regularity and representational fidelity.\textbf{ We prove that the 3:4 format is an optimal for the LUT-based engine in Sec.~\ref{sec: ideal}.}

\subsection{Arenas: Annealing Residual Synapse}
Despite the hardware and bit-utilization advantages of 3:4 structured sparsity, its direct application in ternary quantization often triggers significant performance degradation compared to the dense model. We identify the root cause as \textbf{weight trapping}. As illustrated in Fig.~\ref{fig: problem}, under the hard 3:4 pruning constraint, weights tend to polarize toward specific values, resulting in a distribution that mimics binary quantization. This collapse prevents the model from exploiting the expressive capacity of the ternary set, essentially trapping it in a suboptimal binary-like state in Fig.~\ref{fig: more_arenas} \textit{(Top Right)}.

We find that this stagnation is driven by \textbf{Gradient Homogenization}. In a 3:4 sparsity configuration, the zeros are distributed uniformly within $T$, causing the sparse matrix to behave similarly to a dense binary matrix. This uniform distribution mimics the properties of a Hadamard transform~\cite{huang2024rolora, ashkboos2024quarot}, which flattens the signal. Consequently, the downstream gradients w.r.t. the activations $X$ become increasingly homogenized:
\begin{equation}
    \frac{\partial L}{\partial X} = \frac{\partial L}{\partial Y}(T\alpha)^\top.
\end{equation}
Therefore, the gradients passed to preceding layers become undifferentiated. This forces the majority of weights to exhibit similar behavior during training, significantly decreasing the representational diversity. In Fig.~\ref{fig: er} \textit{(left)}, the Effective Rank (ER)~\cite{roy2007effective} for 3:4 sparse training confirms our analysis. Specifically, the 3:4 sparse regime exhibits a low ER ($ER < 750$), a level of spectral collapse comparable to that of binary quantization, despite the gradient matrix having a total dimensionality of 4096. This indicates a significant loss of learning degrees of freedom in naive 3:4 ternary sparse training (see Appendix~\ref{sec: er} for more details).

To restore representational diversity, we propose Arenas, an Annealing Residual Synapse mechanism that re-couples latent weight magnitudes to the loss objective via a continuous bypass. During the training phase, the output of a ternary linear layer is augmented with a decaying full-precision residual synapse:
\begin{equation}
    Y = XT \alpha + \lambda_tXW,
\end{equation}
where  $\lambda_t$ is a scheduling coefficient that anneals to zero by the conclusion of training. The inclusion of the latent matrix $W$ in the forward pass fundamentally alters the gradient dynamics. The gradient with respect to the latent activation $X$ becomes:
\begin{equation}
    \frac{\partial L}{\partial X} = \frac{\partial L}{\partial Y} (T \alpha + \lambda_t W)^\top.
\end{equation}
By incorporating the continuous values of $W$ into the backward path, Arenas injects heterogeneous information back into $\frac{\partial L}{\partial X}$, breaking the homogenization effect induced by the 3:4 structure. This allows earlier layers to receive specialized updates, providing the "energy" required for weights to escape trapped binary-like states. As $\lambda_t \to 0$, the residual vanishes, leaving a pure 3:4 sparse ternary model for inference with zero additional overhead.
\begin{table*}[!t]
\centering
\begin{tabular}{c|lc|cccccc}
\hline
 Size & Method  & Bit-width & ARC-e & ARC-c & HelS & PIQA & WinG & Average \\ \hline
\multirow{8}{*}{1B} & {\color{gray}BF16} & {\color{gray}16} & {\color{gray} 0.654} &  {\color{gray} 0.313} & {\color{gray} 0.477} & {\color{gray} 0.742} & {\color{gray} 0.603} & {\color{gray} 0.558}  \\ 
 & LSQ & 1.67 & 0.376 & 0.177 & 0.258 & 0.574 & 0.506  & 0.378 \\
 & SEQ &  1.67 &0.421 & 0.180 & 0.273 & 0.604 & 0.510  & 0.398 \\
 & DLT & 1.67 & 0.424 & 0.174 & 0.256 & 0.563 & 0.513  & 0.386 \\
 & TWN &  1.67 &0.407 & 0.220 & 0.284 & 0.601 & 0.492 & 0.401 \\
 & AbsMedian & 1.67 & 0.567 &  0.251 & 0.339 & 0.674  & 0.533 & 0.473 \\
 & AbsMean &  1.67 &0.603 & 0.259 & 0.360 & 0.683 & 0.541 & {\color{myblue} \textbf{0.489}} \\
 & Tequila &  1.67 & {\color{myblue} \textbf{0.645}} & {\color{myblue} \textbf{0.305}} & {\color{mypurple} \textbf{0.391}} & {\color{mypurple} \textbf{0.710}} & {\color{myblue} \textbf{0.542}} & {\color{mypurple} \textbf{0.519}} \\ 	
  &  {\color{mypurple}\textbf{Sherry}} &  {\color{mypurple}\textbf{1.25}} & {\color{mypurple} \textbf{0.647}} & {\color{mypurple} \textbf{0.309}} & {\color{myblue} \textbf{0.388}} & {\color{myblue} \textbf{0.699}} & {\color{mypurple} \textbf{0.550}} & {\color{mypurple} \textbf{0.519}} \\  \hline	
\multirow{8}{*}{3B} & {\color{gray}BF16} &  16 & {\color{gray} 0.745} &{\color{gray} 0.422} & {\color{gray} 0.552} & {\color{gray}0.768} & {\color{gray}0.691} & {\color{gray}0.636} \\ 
 & LSQ &  1.67 &0.431  & 0.200  & 0.294  & 0.599 & 0.522  & 0.409 \\
 & SEQ &  1.67 &0.498  & 0.231  & 0.303  & 0.645  & 0.529 & 0.441 \\
 & DLT &  1.67 &0.361  & 0.161 & 0.260  & 0.572  & 0.496   & 0.370  \\
 & TWN & 1.67 & 0.692 & {\color{myblue} \textbf{0.351}} & {\color{myblue}\textbf{0.462}} & 0.734 & 0.586  & 0.565 \\
 & AbsMedian & 1.67 & 0.636  & 0.299  & 0.406 & 0.713 & 0.558   & 0.522  \\
 & AbsMean &  1.67 & 0.672  & 0.329  & 0.439 & 0.735  & 0.582  & 0.551  \\
 & Tequila & 1.67 &{\color{mypurple} \textbf{0.702}}  & 0.346  & {\color{mypurple} \textbf{0.464}}  & {\color{mypurple} \textbf{0.739}}  & {\color{mypurple} \textbf{0.627}}  & {\color{mypurple} \textbf{0.576}} \\
  &  {\color{mypurple}\textbf{Sherry}} &  {\color{mypurple}\textbf{1.25}} & {\color{myblue} \textbf{0.688}} & {\color{mypurple} \textbf{0.364}} &  0.452 & {\color{myblue} \textbf{0.736}} & {\color{myblue} \textbf{0.593}} & {\color{myblue} \textbf{0.567}} \\  \hline
\end{tabular}
\vspace{-0.5em}
\caption{Comparison of Sherry with different ternary quantization methods.}
\vspace{-1.5em}
\label{tb: main_result}
\end{table*}
The Arenas provides three critical advantages for 3:4 sparse ternary training:\\
\textbf{(1) Variance Injection and Singularity Breaking:} By re-introducing the continuous matrix $W$, Arenas prevents the gradients $\frac{\partial L}{\partial X}$ from collapsing into a homogenized, low-rank state, allowing earlier layers to receive specialized updates (in Fig.~\ref{fig: er}). \\
\textbf{(2) Adaptive Error Compensation:} During training, the residual term $\lambda_t XW$ naturally absorbs the quantization noise and pruning error introduced by the 3:4 ternary constraint. This allows the network to maintain a high-precision internal representation while the sparse ternary weights ($T\alpha$) gradually learn to capture the most salient components of the signal, leading to better optimization. \\
\textbf{(3) Zero-Overhead Inference:} Because $\lambda_t$ anneals to zero, the auxiliary path is completely removed post-training,  without any inference overheads.

\section{Evaluation}
To validate the efficacy of \method, we conduct experiments evaluating its performance against ternary quantization baselines. All results are averaged over three independent runs with random seeds. In all tables, the best and second-best results are highlighted in {\color{mypurple}\textbf{purple}} and {\color{myblue}\textbf{blue}}; half-precision (BF16) results are shown in {\color{gray}gray} for reference.

\subsection{Experimental Setup}
We provide an overview of our experimental configuration below; further implementation details are available in Appendix~\ref{sec: setup}.

\paragraph{Datasets, Models, and Evaluation:} We utilize LLaMA-3.2-1B and LLaMA-3.2-3B~\citep{touvron2023llama} as our base architectures. Unless otherwise specified, we employ group-wise quantization with group size 128. For quantization-aware training, we utilize 10B tokens sampled from the UltraFineWeb dataset~\citep{wang2025ultra}. Following established ternary quantization benchmarks~\citep{liu2025paretoq, chen2024ternaryllm, ma2025bitnet}, we evaluate performance using \texttt{lm-evaluation-harness}~\citep{eval-harness} across five zero-shot tasks: PIQA~\citep{bisk2020piqa}, ARC-Easy (ARC-e), ARC-Challenge (ARC-c)~\citep{clark2018think}, HellaSwag (HelS)~\citep{zellers2019hellaswag} and WinoGrande (WinG)~\citep{sakaguchi2021winogrande}. In our results, we report a bit-width of $1.67$ for existing ternary quantization baselines rather than the theoretical $1.58$, as $1.67$ reflects the actual bit-density achieved by current practical packing strategies.

\paragraph{Ternary Quantization Baselines:} We compare \method against two categories of quantization methods: (1) \textit{Static methods}, including TWN~\cite{li2016ternary}, Tequila used in TequilaLLM~\cite{huang2025tequila}, and the AbsMedian/AbsMean strategies utilized in BitNet~\citep{ma2025bitnet, wang2023bitnet} and Spectra~\citep{kaushal2025surprising}; and (2) \textit{Learnable methods}, such as DLT~\citep{chen2024ternaryllm} used in TernaryLLM, LSQ~\citep{esser2019learned}, and the SEQ strategy used in ParetoQ~\citep{liu2025paretoq}.

\paragraph{Implementation Details:} Training is conducted on 32 NVIDIA GPUs, while inference throughput is evaluated on an Intel 8263C CPU to verify edge-efficiency. We quantize all linear layers within the Transformer architecture using a sequence length of 1024. The learning rate is fixed at $10^{-4}$.  The annealing coefficient $\lambda_t$ follows a cosine-decay schedule with warmup by default, as shown in Fig.~\ref{fig: schedule}.

\begin{table*}[!t]
\centering
\begin{tabular}{l|ll|lllllll}
\hline
 Model  & Size & Bit-width & ARC-e & ARC-c & HelS & PIQA & WinG & Average\\ \hline
 {\color{gray}LLaMA3.2} & {\color{gray}1B}  & {\color{gray}16} & {\color{gray} 0.654} &  {\color{gray} 0.313} & {\color{gray} 0.477} & {\color{gray} 0.742}  & {\color{gray} 0.603} & {\color{gray} 0.558}\\ 
 TernaryLLM$^*$ & 1B  & 1.67 & 0.424 & 0.174 & 0.256 & 0.563 & 0.513 & 0.386\\
 ParetoQ$^*$ & 1B & 1.67 & 0.421& 0.180& 0.273 &0.604 &0.510& 0.398\\
 LLM-QAT & 1B & 1.67 & 0.360 &  0.262& 0.313& 0.551 & 0.496& 0.397\\
 BitNet & 1.3B &  1.67 & 0.549 & 0.242 & 0.377 & 0.688 &  {\color{mypurple} \textbf{0.558}} & 0.483\\
 Spectra & 1.1B & 1.67 & 0.563 & 0.246 &  {\color{myblue} \textbf{0.388}} &  0.693& {\color{myblue} \textbf{0.555}}& {\color{myblue} \textbf{0.489}}\\
 TequilaLLM & 1B & 1.67 &  {\color{myblue} \textbf{0.645}} & {\color{myblue} \textbf{0.305}} & {\color{mypurple} \textbf{0.391}} & {\color{mypurple} \textbf{0.710}} & 0.542 & {\color{mypurple} \textbf{0.519}} \\ 
  {\color{mypurple}\textbf{SherryLLM}} & 1B &  {\color{mypurple}\textbf{1.25}} & {\color{mypurple} \textbf{0.647}} & {\color{mypurple} \textbf{0.309}} & {\color{myblue} \textbf{0.388}} & {\color{myblue} \textbf{0.699}} & 0.550 & {\color{mypurple} \textbf{0.519}} \\ \hline
{\color{gray}LLaMA3.2} & {\color{gray}3B}  & {\color{gray}16} &  {\color{gray}0.745} &  {\color{gray}0.422} &  {\color{gray}0.552}&  {\color{gray}0.768} &  {\color{gray}0.691} &   {\color{gray}0.636} \\ 
TernaryLLM$^*$ & 3B & 1.67  & 0.361 & 0.161 & 0.260  &  0.572  &  0.496 & 0.370 \\
ParetoQ$^*$ & 3B & 1.67 & 0.498  & 0.231  & 0.303  & 0.645  & 0.529& 0.441\\
LLM-QAT & 3B & 1.67 &0.445 &0.307&0.434&0.627&0.506& 0.464\\
BitNet & 3B & 1.67 & 0.614&0.283&0.429&0.715&0.593&0.527 \\
Spectra & 3.9B & 1.67 &0.660& 0.319 &{\color{mypurple} \textbf{0.483}}&{\color{mypurple} \textbf{0.744}}&{\color{mypurple} \textbf{0.631}}& {\color{myblue} \textbf{0.567}}\\
TequilaLLM & 3B & 1.67 & {\color{mypurple}\textbf{0.702}}  & {\color{myblue}\textbf{0.346}}  & {\color{myblue}\textbf{0.464}}  & {\color{myblue}\textbf{0.739}}  & {\color{myblue} \textbf{0.627}}  & {\color{mypurple}\textbf{0.576}} \\
{\color{mypurple} \textbf{SherryLLM}} & 3B & {\color{mypurple}\textbf{1.25}}&  {\color{myblue} \textbf{0.688}} & {\color{mypurple} \textbf{0.364}} &  0.452 & 0.736 & 0.593 & {\color{myblue} \textbf{0.567}}\\
 \hline
\end{tabular}
\vspace{-0.5em}
\caption{Comparison of SherryLLM with LLaMA-based ternary LLMs across different model sizes; $^*$ indicates results obtained from our reproduction.}
\vspace{-1.5em}
\label{tb: main_result_2}
\end{table*}

\begin{figure}[t]
    \centering    
 \includegraphics[width=0.9\linewidth]{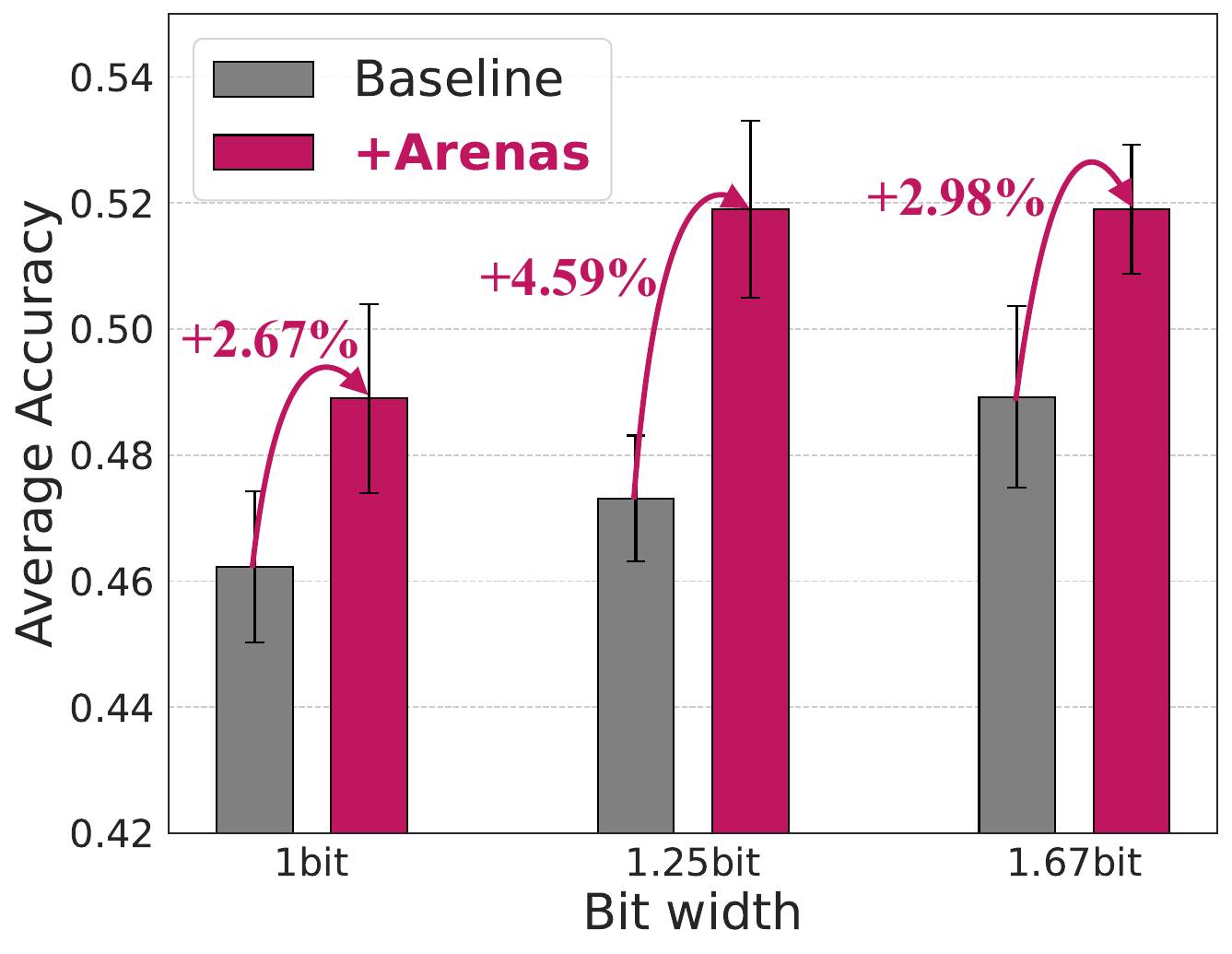}
    \vspace{-1em}
    \caption{Ablation study of Arenas.}
    \label{fig: ablation}
    \vspace{-1.5em}
\end{figure}

\subsection{Performance Evaluation}
\paragraph{Comparison of Ternary Quantization Methods:} As shown in Table~\ref{tb: main_result}, \textbf{Sherry} achieves performance parity with the current SOTA, Tequila, across both 1B and 3B scales, despite utilizing a significantly lower bit-width of 1.25 bits. On the 1B model, Sherry matches the average SOTA accuracy exactly, effectively matching the 1.67-bit baseline while achieving a 25\% reduction in bit-width. Notably, on reasoning-intensive benchmarks like ARC-Challenge, Sherry even outperforms Tequila and narrows the gap to the full-precision BF16 baseline to less than 0.5\%. These results demonstrate that 3:4 structured sparsity preserves high-level linguistic capabilities, while the Arenas module successfully addresses optimization trapping. This synergy enables a high-efficiency packing strategy that simultaneously ensures superior hardware alignment and competitive model quality.

\paragraph{Comparison of Ternary LLMs:} To further evaluate the effectiveness of Sherry, we denote the resulting models as \textbf{SherryLLM} and compare them against existing ternary LLaMA-based architectures. For a fair comparison, we reproduced methods with available training code~\cite{liu2025paretoq, chen2024ternaryllm} on 10B tokens from the UltraFineWeb dataset using identical hyperparameters, while reporting original publication results for others. As shown in Table~\ref{tb: main_result_2}, \textbf{SherryLLM} achieves at least second-best average accuracy across both scales despite its significantly lower bit-width of 1.25 bits. These results confirm that the integration of 3:4 structured sparsity with the Arenas module preserves model expressivity, allowing SherryLLM to achieve a $25\%$ bit-width reduction while maintaining competitive performance against baselines.
\begin{table}[t]
    \centering
    \begin{tabular}{c|c}
        \hline
        Granularity & Average Acc $\pm$ Std \\
        \hline
        Per-tensor & 0.502 $\pm$  0.010\\
        Per-channel & 0.513 $\pm$ 0.011\\
        Per-group & 0.519 $\pm$ 0.014\\
        \hline
    \end{tabular}
    \vspace{-0.5em}
    \caption{Average accuracy of Sherry across various quantization granularities.}
    \vspace{-1.5em}
    \label{tb: gran}
\end{table}

\begin{figure*}[t]
\begin{minipage}[t]{0.48\textwidth}
    \centering    
    \includegraphics[width=0.9\linewidth]{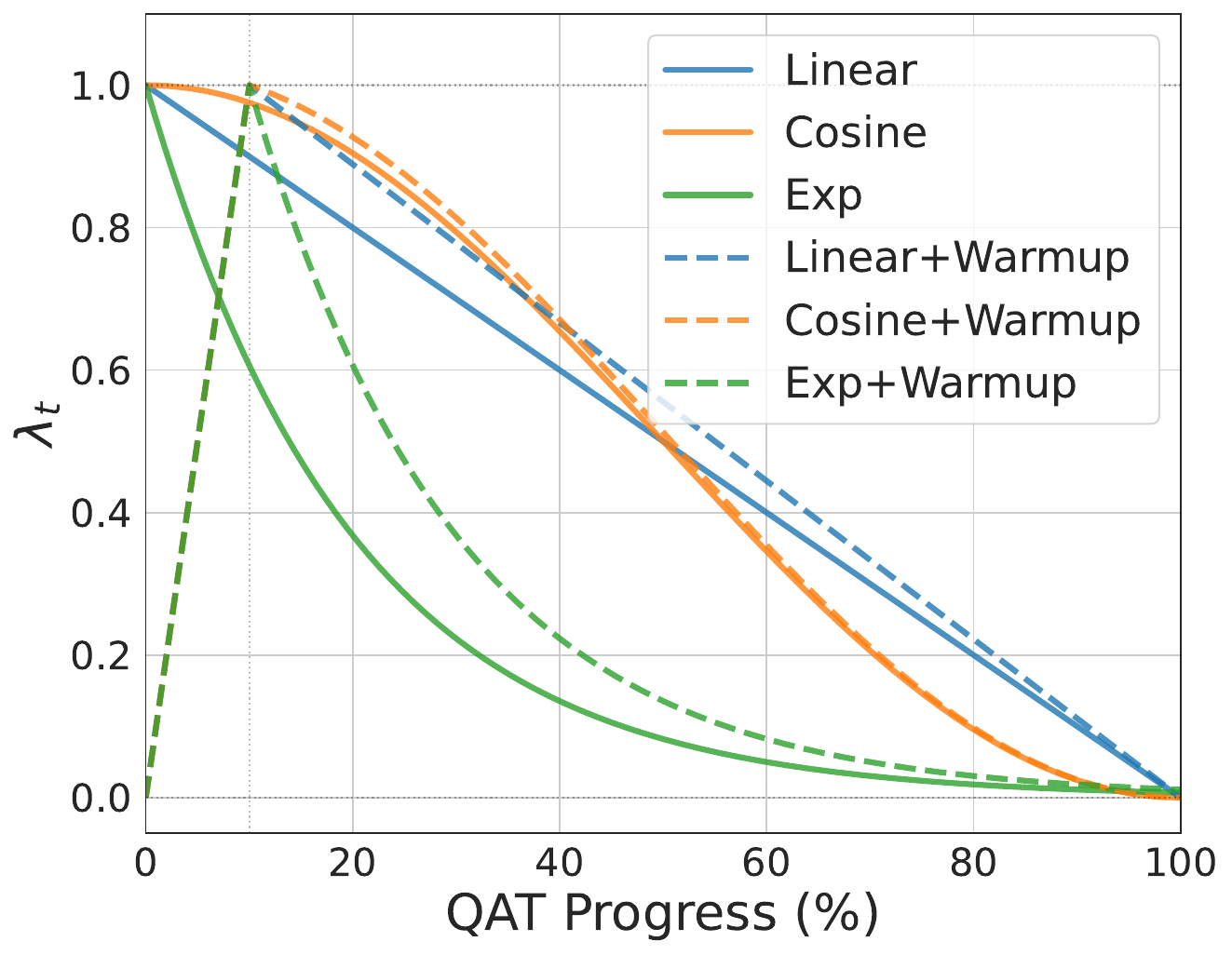}
    \vspace{-1em}
    \caption{The schedules of annealing factor $\lambda_t$.}
    \label{fig: schedule}
    \vspace{-1.5em}
\end{minipage}
\hspace{0.2em}
\begin{minipage}[t]{0.48\textwidth}
    \centering    
    \includegraphics[width=0.9\linewidth]{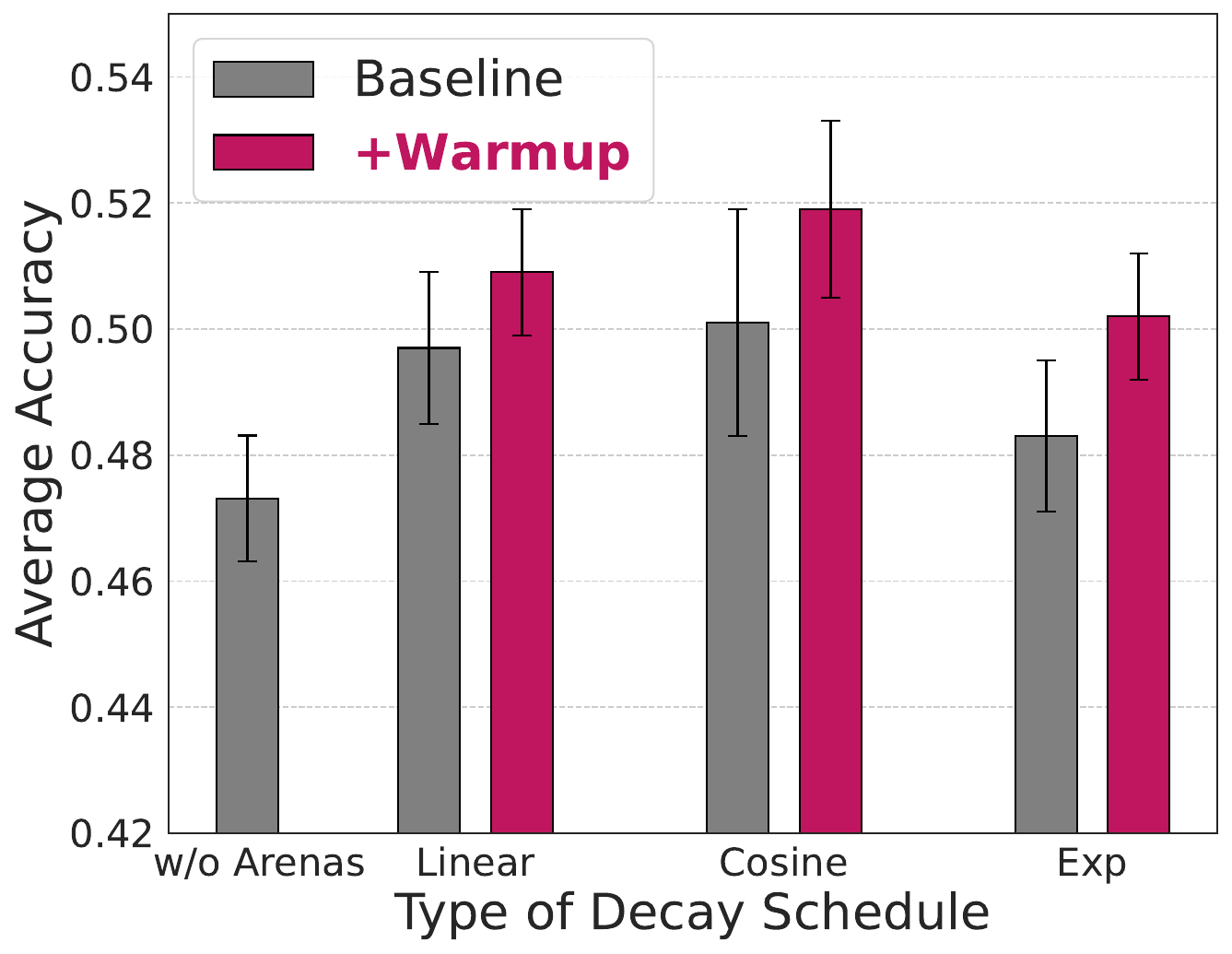}
    \vspace{-1em}
    \caption{Comparison of different schedules for $\lambda_t$. }
    \label{fig: schedule_2}
    \vspace{-1.5em}
\end{minipage}
\end{figure*}

\paragraph{Impact of Quantization Granularity:} Quantization granularity involves a critical trade-off between hardware efficiency and representational accuracy. While coarse-grained per-tensor quantization maximizes acceleration, it often introduces significant error, whereas fine-grained per-group strategies mitigate this at the cost of increased memory overhead for scaling factors. We evaluate Sherry across per-tensor, per-channel, and per-group (size 128) granularities on five benchmarks. As the results of average accuracy shown in Table~\ref{tb: gran}, Sherry maintains robust performance across all levels with minimal degradation. This stability is primarily driven by the Arenas module, which provides a continuous gradient flow that allows latent weights to adapt to varying scaling constraints. 

\paragraph{Effectiveness of Arenas:} To formally demonstrate the effectiveness of our training framework, we conducted an ablation study across three quantization schemes: binary (1-bit), 3:4 structured sparse (1.25-bit), and pure ternary AbsMean (1.67-bit). As shown in the Fig~\ref{fig: ablation}, the integration of Arenas yields consistent performance improvements across all configurations. Notably, all the schemes exhibit significant gains. We attribute this to the Arenas’s ability to mitigate the trapping issue in both 1-bit and 1.67-bit, as visualized in Fig.~\ref{fig: more_arenas}.

\paragraph{Inference Efficiency:}
To empirically validate the efficiency of \method, we evaluate the token generation speed against 1.67-bit (TL2) and 2-bit (I2\_S) baselines within the BitNet.cpp framework~\citep{wang2025bitnetcpp} using per-channel quantization, alongside a BF16 baseline. Experiments were conducted on an Intel i7-14700HX CPU using 700M and 3B BitNet variants. Both \method and the baselines utilize the BitNet.cpp paradigm. The results in Table~\ref{tab:inference_speed} demonstrate that \method outperforms 1.67-bit and 2-bit variants. Specifically, for the 3B model, \method achieves a $18\%$ speedup over the 1.67-bit baseline. This improvement is attributed to our 3:4 structured sparsity and hardware-aligned 5-bit packing, which maximizes SIMD vector lane utilization and eliminates the bit-shuffling overhead inherent in non-power-of-two packing schemes.

\begin{table}[t]
\centering
\begin{tabular}{l|cc|cc}\hline
Scale &Method & Bits & Speed (t/s)$\uparrow$  &Size(MB)$\downarrow$ \\ \hline
\multirow{4}{*}{0.7B} & \color{gray}BF16 & \color{gray}16 & \color{gray}34.01 & \color{gray}1360.0 \\
 & I2\_S & 2.0 & {\color{myblue}\textbf{132.13}} & 256.56 \\
& TL2 & 1.67 & 116.83 & {\color{myblue}\textbf{233.44}} \\
& {\color{mypurple}\textbf{Sherry}} & {\color{mypurple}\textbf{1.25}} & {\color{mypurple}\textbf{148.27}} & {\color{mypurple}\textbf{205.50}} \\ \hline
\multirow{4}{*}{3B}& \color{gray}BF16 & \color{gray}16 & \color{gray}7.55 & \color{gray}6190.0 \\
 & I2\_S & 2.0 & {\color{myblue}\textbf{41.87}} & 873.65 \\
 & TL2 & 1.67 & 38.80 & {\color{myblue}\textbf{846.01}} \\
 & {\color{mypurple}\textbf{Sherry}} & {\color{mypurple}\textbf{1.25}} & {\color{mypurple}\textbf{45.55}} & {\color{mypurple}\textbf{712.40}} \\ \hline
\end{tabular}
\vspace{-0.5em}
\caption{Inference efficiency on Intel i7-14700HX. }
\vspace{-1.5em}
\label{tab:inference_speed}
\end{table}

\paragraph{Impact of $\lambda_t$ Schedules:} To evaluate the sensitivity of the annealing process, we compare three decay schedules for the annealing gate $\lambda_t$: linear, exponential, and cosine, alongside their warmup counterparts. As shown in Fig.~\ref{fig: schedule_2}, every schedule consistently outperforms the baseline without Arenas, demonstrating the robust effectiveness of the Arenas regardless of the specific decay curve. Notably, warmup can increase performance for all schedules. We attribute it from early-stage optimization dynamics in warmup: by gradually introducing the residual, the model establishes a stable foundation before the annealing influence peaks. 

\section{Conclusion}
In this work, we introduce \method, a novel ternary quantization framework designed to resolve the trade-off between bit width and inference speed. By leveraging a 3:4 structured sparsity pattern, Sherry achieves a 1.25-bit-width that natively aligns with SIMD vector lanes, effectively resolving the hardware under-utilization issues common in previous ternary packing strategies. To combat the performance degradation due to weight trapping, we proposed the Arenas module. Arenas provides an annealing, heterogeneous gradient flow during the training phase, allowing sparse models to retain the expressive diversity. Our extensive evaluations on the LLaMA-3.2 models demonstrate that \method achieves performance parity with SOTA ternary models while utilizing 25\% fewer bits and $10\%$ speed up.

\section*{Limitation}
While \method advances the efficiency of edge-deployed LLMs, several limitations offer avenues for future research:

\paragraph{Edge-Centric Model Scale:} Our evaluation focuses on models up to 3B parameters, as these are the primary candidates for local edge deployment. While we demonstrate that \method achieves a superior Pareto frontier for these scales, the behavior of the Arenas mechanism and the 3:4 sparsity pattern on larger, server-grade models (70B+) remains to be validated.

\paragraph{Lack of Server-Specific Optimizations:} To prioritize edge inference efficiency, our experiments focus on general SIMD-aligned vector packing rather than server-specific hardware optimizations, such as NVIDIA Sparse Tensor Cores. While this does not affect Sherry's utility for mobile and local processors, future benchmarks on data-center GPUs could further explore its potential for high-throughput server applications.

\paragraph{Weight-Only Quantization:}To maximize weight-streaming throughput on edge devices, this study focuses on weight-only ternary quantization. While 1.25-bit weights drastically reduce the static memory footprint, the activations and KV-cache remain in BF16. Future integration with activation quantization could further alleviate memory bottlenecks during long-context inference.

\paragraph{Training Overhead:}\method is designed for maximum inference-phase efficiency. However, the Arenas mechanism increases the compute overhead during the training (QAT) phase. While this is a one-time cost that does not affect edge deployment, it is a factor to consider for researchers with restricted training budgets.

\bibliography{main}

\begin{thebibliography}{60}
\providecommand{\natexlab}[1]{#1}

\bibitem[{Ashkboos et~al.(2024)Ashkboos, Mohtashami, Croci, Li, Cameron, Jaggi, Alistarh, Hoefler, and Hensman}]{ashkboos2024quarot}
Saleh Ashkboos, Amirkeivan Mohtashami, Maximilian~L Croci, Bo~Li, Pashmina Cameron, Martin Jaggi, Dan Alistarh, Torsten Hoefler, and James Hensman. 2024.
\newblock Quarot: Outlier-free 4-bit inference in rotated llms.
\newblock \emph{arXiv preprint arXiv:2404.00456}.

\bibitem[{Bisk et~al.(2020)Bisk, Zellers, Gao, Choi et~al.}]{bisk2020piqa}
Yonatan Bisk, Rowan Zellers, Jianfeng Gao, Yejin Choi, and 1 others. 2020.
\newblock Piqa: Reasoning about physical commonsense in natural language.
\newblock In \emph{Proceedings of the AAAI conference on artificial intelligence}, volume~34, pages 7432--7439.

\bibitem[{Chen et~al.(2025)Chen, Shao, Xu, Wang, Gao, Zhang, and Luo}]{chen2025efficientqat}
Mengzhao Chen, Wenqi Shao, Peng Xu, Jiahao Wang, Peng Gao, Kaipeng Zhang, and Ping Luo. 2025.
\newblock Efficientqat: Efficient quantization-aware training for large language models.
\newblock In \emph{Proceedings of the 63rd Annual Meeting of the Association for Computational Linguistics (Volume 1: Long Papers)}, pages 10081--10100.

\bibitem[{Chen et~al.(2024{\natexlab{a}})Chen, Qiu, Zhou, Zhang, Si, and Wu}]{chen2024distributed}
Ning Chen, Tie Qiu, Xiaobo Zhou, Songwei Zhang, Weisheng Si, and Dapeng~Oliver Wu. 2024{\natexlab{a}}.
\newblock A distributed co-evolutionary optimization method with motif for large-scale iot robustness.
\newblock \emph{IEEE/ACM Transactions on Networking}.

\bibitem[{Chen et~al.(2024{\natexlab{b}})Chen, Li, Xu, Zhu, Li, Tian, Barsoum, Wang, and Cheng}]{chen2024ternaryllm}
Tianqi Chen, Zhe Li, Weixiang Xu, Zeyu Zhu, Dong Li, Lu~Tian, Emad Barsoum, Peisong Wang, and Jian Cheng. 2024{\natexlab{b}}.
\newblock Ternaryllm: Ternarized large language model.
\newblock \emph{arXiv preprint arXiv:2406.07177}.

\bibitem[{Clark et~al.(2018)Clark, Cowhey, Etzioni, Khot, Sabharwal, Schoenick, and Tafjord}]{clark2018think}
Peter Clark, Isaac Cowhey, Oren Etzioni, Tushar Khot, Ashish Sabharwal, Carissa Schoenick, and Oyvind Tafjord. 2018.
\newblock Think you have solved question answering? try arc, the ai2 reasoning challenge.
\newblock \emph{arXiv preprint arXiv:1803.05457}.

\bibitem[{Dettmers et~al.(2022)Dettmers, Lewis, Belkada, and Zettlemoyer}]{dettmers2022llm}
Tim Dettmers, Mike Lewis, Younes Belkada, and Luke Zettlemoyer. 2022.
\newblock Llm. int8 (): 8-bit matrix multiplication for transformers at scale.
\newblock \emph{arXiv preprint arXiv:2208.07339}.

\bibitem[{Dettmers et~al.(2021)Dettmers, Lewis, Shleifer, and Zettlemoyer}]{dettmers20218}
Tim Dettmers, Mike Lewis, Sam Shleifer, and Luke Zettlemoyer. 2021.
\newblock 8-bit optimizers via block-wise quantization.
\newblock \emph{arXiv preprint arXiv:2110.02861}.

\bibitem[{Dettmers et~al.(2024)Dettmers, Pagnoni, Holtzman, and Zettlemoyer}]{dettmers2024qlora}
Tim Dettmers, Artidoro Pagnoni, Ari Holtzman, and Luke Zettlemoyer. 2024.
\newblock Qlora: Efficient finetuning of quantized llms.
\newblock \emph{Advances in Neural Information Processing Systems}, 36.

\bibitem[{Esser et~al.(2019)Esser, McKinstry, Bablani, Appuswamy, and Modha}]{esser2019learned}
Steven~K Esser, Jeffrey~L McKinstry, Deepika Bablani, Rathinakumar Appuswamy, and Dharmendra~S Modha. 2019.
\newblock Learned step size quantization.
\newblock \emph{arXiv preprint arXiv:1902.08153}.

\bibitem[{Floridi and Chiriatti(2020)}]{floridi2020gpt}
Luciano Floridi and Massimo Chiriatti. 2020.
\newblock Gpt-3: Its nature, scope, limits, and consequences.
\newblock \emph{Minds and Machines}, 30:681--694.

\bibitem[{Forman and Zahorjan(1994)}]{forman1994challenges}
George~H. Forman and John Zahorjan. 1994.
\newblock The challenges of mobile computing.
\newblock \emph{Computer}, 27(4):38--47.

\bibitem[{Frantar et~al.(2022)Frantar, Ashkboos, Hoefler, and Alistarh}]{frantar2022gptq}
Elias Frantar, Saleh Ashkboos, Torsten Hoefler, and Dan Alistarh. 2022.
\newblock Gptq: Accurate post-training quantization for generative pre-trained transformers.
\newblock \emph{arXiv preprint arXiv:2210.17323}.

\bibitem[{Fu et~al.(2023)Fu, Yang, So, Lam, Bing, and Collier}]{fu2023effectiveness}
Zihao Fu, Haoran Yang, Anthony Man-Cho So, Wai Lam, Lidong Bing, and Nigel Collier. 2023.
\newblock On the effectiveness of parameter-efficient fine-tuning.
\newblock In \emph{Proceedings of the AAAI conference on artificial intelligence}, volume~37, pages 12799--12807.

\bibitem[{Gao et~al.(2024)Gao, Tow, Abbasi, Biderman, Black, DiPofi, Foster, Golding, Hsu, Le~Noac'h, Li, McDonell, Muennighoff, Ociepa, Phang, Reynolds, Schoelkopf, Skowron, Sutawika, Tang, Thite, Wang, Wang, and Zou}]{eval-harness}
Leo Gao, Jonathan Tow, Baber Abbasi, Stella Biderman, Sid Black, Anthony DiPofi, Charles Foster, Laurence Golding, Jeffrey Hsu, Alain Le~Noac'h, Haonan Li, Kyle McDonell, Niklas Muennighoff, Chris Ociepa, Jason Phang, Laria Reynolds, Hailey Schoelkopf, Aviya Skowron, Lintang Sutawika, and 5 others. 2024.
\newblock \href {https://doi.org/10.5281/zenodo.12608602} {The language model evaluation harness}.

\bibitem[{Huang and Wu(2025)}]{huang2025quaff}
Hong Huang and Dapeng Wu. 2025.
\newblock Quaff: Quantized parameter-efficient fine-tuning under outlier spatial stability hypothesis.
\newblock \emph{arXiv preprint arXiv:2505.14742}.

\bibitem[{Huang et~al.(2025{\natexlab{a}})Huang, Wu, Cen, Yu, Li, Liu, Zhu, Chen, Liu, and Wu}]{huang2025tequila}
Hong Huang, Decheng Wu, Rui Cen, Guanghua Yu, Zonghang Li, Kai Liu, Jianchen Zhu, Peng Chen, Xue Liu, and Dapeng Wu. 2025{\natexlab{a}}.
\newblock Tequila: Trapping-free ternary quantization for large language models.
\newblock \emph{arXiv preprint arXiv:2509.23809}.

\bibitem[{Huang et~al.(2025{\natexlab{b}})Huang, Yang, Chen, Ye, and Wu}]{huang2025fedrts}
Hong Huang, Hai Yang, Yuan Chen, Jiaxun Ye, and Dapeng Wu. 2025{\natexlab{b}}.
\newblock Fedrts: Federated robust pruning via combinatorial thompson sampling.
\newblock \emph{arXiv preprint arXiv:2501.19122}.

\bibitem[{Huang et~al.(2023)Huang, Zhang, Sun, Fang, Yuan, and Wu}]{huang2023distributed}
Hong Huang, Lan Zhang, Chaoyue Sun, Ruogu Fang, Xiaoyong Yuan, and Dapeng Wu. 2023.
\newblock Distributed pruning towards tiny neural networks in federated learning.
\newblock In \emph{2023 IEEE 43rd International Conference on Distributed Computing Systems (ICDCS)}, pages 190--201. IEEE.

\bibitem[{Huang et~al.(2024{\natexlab{a}})Huang, Zhuang, Chen, and Lyu}]{huang2024fedmef}
Hong Huang, Weiming Zhuang, Chen Chen, and Lingjuan Lyu. 2024{\natexlab{a}}.
\newblock Fedmef: Towards memory-efficient federated dynamic pruning.
\newblock In \emph{Proceedings of the IEEE/CVF Conference on Computer Vision and Pattern Recognition}, pages 27548--27557.

\bibitem[{Huang et~al.(2024{\natexlab{b}})Huang, Liu, Liu, and Cheng}]{huang2024rolora}
Xijie Huang, Zechun Liu, Shih-Yang Liu, and Kwang-Ting Cheng. 2024{\natexlab{b}}.
\newblock Rolora: Fine-tuning rotated outlier-free llms for effective weight-activation quantization.
\newblock \emph{arXiv preprint arXiv:2407.08044}.

\bibitem[{Huang et~al.(2025{\natexlab{c}})Huang, Ma, Cao, Shu, Wang, Cao, Chen, and Xiong}]{huang2025tenet}
Zhirui Huang, Rui Ma, Shijie Cao, Ran Shu, Ian Wang, Ting Cao, Chixiao Chen, and Yongqiang Xiong. 2025{\natexlab{c}}.
\newblock Tenet: An efficient sparsity-aware lut-centric architecture for ternary llm inference on edge.
\newblock \emph{arXiv preprint arXiv:2509.13765}.

\bibitem[{Hubara et~al.(2018)Hubara, Courbariaux, Soudry, El-Yaniv, and Bengio}]{hubara2018quantized}
Itay Hubara, Matthieu Courbariaux, Daniel Soudry, Ran El-Yaniv, and Yoshua Bengio. 2018.
\newblock Quantized neural networks: Training neural networks with low precision weights and activations.
\newblock \emph{Journal of Machine Learning Research}, 18(187):1--30.

\bibitem[{Imielinski and Korth(1996)}]{imielinski1996mobile}
Tomasz Imielinski and Henry~F Korth. 1996.
\newblock \emph{Mobile computing}, volume 353.
\newblock Springer Science \& Business Media.

\bibitem[{Kaushal et~al.(2025)Kaushal, Vaidhya, Mondal, Pandey, Bhagat, and Rish}]{kaushal2025surprising}
Ayush Kaushal, Tejas Vaidhya, Arnab~Kumar Mondal, Tejas Pandey, Aaryan Bhagat, and Irina Rish. 2025.
\newblock Surprising effectiveness of pretraining ternary language model at scale.
\newblock In \emph{The Thirteenth International Conference on Learning Representations}.

\bibitem[{Leng et~al.(2018)Leng, Dou, Li, Zhu, and Jin}]{leng2018extremely}
Cong Leng, Zesheng Dou, Hao Li, Shenghuo Zhu, and Rong Jin. 2018.
\newblock Extremely low bit neural network: Squeeze the last bit out with admm.
\newblock In \emph{Proceedings of the AAAI conference on artificial intelligence}, volume~32.

\bibitem[{Li et~al.(2016)Li, Liu, Wang, Zhang, and Yan}]{li2016ternary}
Fengfu Li, Bin Liu, Xiaoxing Wang, Bo~Zhang, and Junchi Yan. 2016.
\newblock Ternary weight networks.
\newblock \emph{arXiv preprint arXiv:1605.04711}.

\bibitem[{Li et~al.(2023)Li, Yu, Liang, He, Karampatziakis, Chen, and Zhao}]{li2023loftq}
Yixiao Li, Yifan Yu, Chen Liang, Pengcheng He, Nikos Karampatziakis, Weizhu Chen, and Tuo Zhao. 2023.
\newblock Loftq: Lora-fine-tuning-aware quantization for large language models.
\newblock \emph{arXiv preprint arXiv:2310.08659}.

\bibitem[{Li et~al.(2025)Li, Li, Feng, Guizani, and Yu}]{li2025prima}
Zonghang Li, Tao Li, Wenjiao Feng, Mohsen Guizani, and Hongfang Yu. 2025.
\newblock Prima. cpp: Speeding up 70b-scale llm inference on low-resource everyday home clusters.
\newblock \emph{arXiv preprint arXiv:2504.08791}.

\bibitem[{Liagkou et~al.(2024)Liagkou, Filiopoulou, Fragiadakis, Nikolaidou, and Michalakelis}]{liagkou2024cost}
Vasiliki Liagkou, Evangelia Filiopoulou, George Fragiadakis, Mara Nikolaidou, and Christos Michalakelis. 2024.
\newblock The cost perspective of adopting large language model-as-a-service.
\newblock In \emph{2024 IEEE International Conference on Joint Cloud Computing (JCC)}, pages 80--83. IEEE.

\bibitem[{Lin et~al.(2023{\natexlab{a}})Lin, Zheng, Wang, Cao, Ma, Zhang, Zhu, Cao, Xue, Yang et~al.}]{lin2023efficient}
Bin Lin, Ningxin Zheng, Lei Wang, Shijie Cao, Lingxiao Ma, Quanlu Zhang, Yi~Zhu, Ting Cao, Jilong Xue, Yuqing Yang, and 1 others. 2023{\natexlab{a}}.
\newblock Efficient gpu kernels for n: M-sparse weights in deep learning.
\newblock \emph{Proceedings of Machine Learning and Systems}, 5:513--525.

\bibitem[{Lin et~al.(2025)Lin, Xu, Wu, Guo, Zhang, Lu, Wei, Zhang, and Sun}]{lin2025quantization}
Haokun Lin, Haobo Xu, Yichen Wu, Ziyu Guo, Renrui Zhang, Zhichao Lu, Ying Wei, Qingfu Zhang, and Zhenan Sun. 2025.
\newblock Quantization meets dllms: A systematic study of post-training quantization for diffusion llms.
\newblock \emph{arXiv preprint arXiv:2508.14896}.

\bibitem[{Lin et~al.(2023{\natexlab{b}})Lin, Tang, Tang, Yang, Dang, and Han}]{lin2023awq}
Ji~Lin, Jiaming Tang, Haotian Tang, Shang Yang, Xingyu Dang, and Song Han. 2023{\natexlab{b}}.
\newblock Awq: Activation-aware weight quantization for llm compression and acceleration.
\newblock \emph{arXiv preprint arXiv:2306.00978}.

\bibitem[{Liu and Liu(2023)}]{liu2023ternary}
Dan Liu and Xue Liu. 2023.
\newblock Ternary quantization: A survey.
\newblock \emph{arXiv preprint arXiv:2303.01505}.

\bibitem[{Liu et~al.(2023)Liu, Gong, Wei, Dong, Cai, and Zhuang}]{liu2023qllm}
Jing Liu, Ruihao Gong, Xiuying Wei, Zhiwei Dong, Jianfei Cai, and Bohan Zhuang. 2023.
\newblock Qllm: Accurate and efficient low-bitwidth quantization for large language models.
\newblock \emph{arXiv preprint arXiv:2310.08041}.

\bibitem[{Liu et~al.(2025)Liu, Zhao, Huang, Chen, Zhang, Zhao, Roy, Jin, Xiong, Shi et~al.}]{liu2025paretoq}
Zechun Liu, Changsheng Zhao, Hanxian Huang, Sijia Chen, Jing Zhang, Jiawei Zhao, Scott Roy, Lisa Jin, Yunyang Xiong, Yangyang Shi, and 1 others. 2025.
\newblock Paretoq: Scaling laws in extremely low-bit llm quantization.
\newblock \emph{arXiv preprint arXiv:2502.02631}.

\bibitem[{Ma et~al.(2025)Ma, Wang, Huang, Zhang, Hu, Song, Xia, and Wei}]{ma2025bitnet}
Shuming Ma, Hongyu Wang, Shaohan Huang, Xingxing Zhang, Ying Hu, Ting Song, Yan Xia, and Furu Wei. 2025.
\newblock Bitnet b1. 58 2b4t technical report.
\newblock \emph{arXiv preprint arXiv:2504.12285}.

\bibitem[{McMahan et~al.(2017)McMahan, Moore, Ramage, Hampson, and y~Arcas}]{mcmahan2017communication}
Brendan McMahan, Eider Moore, Daniel Ramage, Seth Hampson, and Blaise~Aguera y~Arcas. 2017.
\newblock Communication-efficient learning of deep networks from decentralized data.
\newblock In \emph{Artificial intelligence and statistics}, pages 1273--1282. PMLR.

\bibitem[{Nie et~al.(2025)Nie, Dong, Zhang, Xiao, and Sun}]{nie2025elutq}
Xin Nie, Liang Dong, Haicheng Zhang, Jiawang Xiao, and G~Sun. 2025.
\newblock Elutq: Efficient lut-aware quantization for deploying large language models on edge devices.
\newblock \emph{arXiv preprint arXiv:2510.19482}.

\bibitem[{Roy and Vetterli(2007)}]{roy2007effective}
Olivier Roy and Martin Vetterli. 2007.
\newblock The effective rank: A measure of effective dimensionality.
\newblock In \emph{2007 15th European signal processing conference}, pages 606--610. IEEE.

\bibitem[{Sakaguchi et~al.(2021)Sakaguchi, Bras, Bhagavatula, and Choi}]{sakaguchi2021winogrande}
Keisuke Sakaguchi, Ronan~Le Bras, Chandra Bhagavatula, and Yejin Choi. 2021.
\newblock Winogrande: An adversarial winograd schema challenge at scale.
\newblock \emph{Communications of the ACM}, 64(9):99--106.

\bibitem[{Sun et~al.(2021)Sun, Zhou, Stuijk, Wijnhoven, Nelson, Corporaal et~al.}]{sun2021dominosearch}
Wei Sun, Aojun Zhou, Sander Stuijk, Rob Wijnhoven, Andrew~O Nelson, Henk Corporaal, and 1 others. 2021.
\newblock Dominosearch: Find layer-wise fine-grained n: M sparse schemes from dense neural networks.
\newblock \emph{Advances in neural information processing systems}, 34:20721--20732.

\bibitem[{Team et~al.(2025)Team, Xiao, Li, Han, Bai, Cai, Chen, Chen, Cong, Cui et~al.}]{team2025minicpm4}
MiniCPM Team, Chaojun Xiao, Yuxuan Li, Xu~Han, Yuzhuo Bai, Jie Cai, Haotian Chen, Wentong Chen, Xin Cong, Ganqu Cui, and 1 others. 2025.
\newblock Minicpm4: Ultra-efficient llms on end devices.
\newblock \emph{arXiv preprint arXiv:2506.07900}.

\bibitem[{Touvron et~al.(2023)Touvron, Martin, Stone, Albert, Almahairi, Babaei, Bashlykov, Batra, Bhargava, Bhosale et~al.}]{touvron2023llama}
Hugo Touvron, Louis Martin, Kevin Stone, Peter Albert, Amjad Almahairi, Yasmine Babaei, Nikolay Bashlykov, Soumya Batra, Prajjwal Bhargava, Shruti Bhosale, and 1 others. 2023.
\newblock Llama 2: Open foundation and fine-tuned chat models.
\newblock \emph{arXiv preprint arXiv:2307.09288}.

\bibitem[{Wang et~al.(2023)Wang, Ma, Dong, Huang, Wang, Ma, Yang, Wang, Wu, and Wei}]{wang2023bitnet}
Hongyu Wang, Shuming Ma, Li~Dong, Shaohan Huang, Huaijie Wang, Lingxiao Ma, Fan Yang, Ruiping Wang, Yi~Wu, and Furu Wei. 2023.
\newblock Bitnet: Scaling 1-bit transformers for large language models.
\newblock \emph{arXiv preprint arXiv:2310.11453}.

\bibitem[{Wang et~al.(2025{\natexlab{a}})Wang, Ma, Ma, Wang, Wang, Dong, Huang, Wang, Xue, Wang et~al.}]{wang2025bitnet}
Hongyu Wang, Shuming Ma, Lingxiao Ma, Lei Wang, Wenhui Wang, Li~Dong, Shaohan Huang, Huaijie Wang, Jilong Xue, Ruiping Wang, and 1 others. 2025{\natexlab{a}}.
\newblock Bitnet: 1-bit pre-training for large language models.
\newblock \emph{Journal of Machine Learning Research}, 26(125):1--29.

\bibitem[{Wang et~al.(2025{\natexlab{b}})Wang, Zhou, Song, Cao, Xia, Cao, Wei, Ma, Wang, and Wei}]{wang2025bitnetcpp}
Jinheng Wang, Hansong Zhou, Ting Song, Shijie Cao, Yan Xia, Ting Cao, Jianyu Wei, Shuming Ma, Hongyu Wang, and Furu Wei. 2025{\natexlab{b}}.
\newblock Bitnet. cpp: Efficient edge inference for ternary llms.
\newblock \emph{arXiv preprint arXiv:2502.11880}.

\bibitem[{Wang et~al.(2025{\natexlab{c}})Wang, Fu, Cai, Tang, Lyu, Fang, Zheng, Zhou, Zeng, Xiao et~al.}]{wang2025ultra}
Yudong Wang, Zixuan Fu, Jie Cai, Peijun Tang, Hongya Lyu, Yewei Fang, Zhi Zheng, Jie Zhou, Guoyang Zeng, Chaojun Xiao, and 1 others. 2025{\natexlab{c}}.
\newblock Ultra-fineweb: Efficient data filtering and verification for high-quality llm training data.
\newblock \emph{arXiv preprint arXiv:2505.05427}.

\bibitem[{Wei et~al.(2025)Wei, Cao, Cao, Ma, Wang, Zhang, and Yang}]{wei2025t}
Jianyu Wei, Shijie Cao, Ting Cao, Lingxiao Ma, Lei Wang, Yanyong Zhang, and Mao Yang. 2025.
\newblock T-mac: Cpu renaissance via table lookup for low-bit llm deployment on edge.
\newblock In \emph{Proceedings of the Twentieth European Conference on Computer Systems}, pages 278--292.

\bibitem[{Wei et~al.(2022)Wei, Zhang, Zhang, Gong, Zhang, Zhang, Yu, and Liu}]{wei2022outlier}
Xiuying Wei, Yunchen Zhang, Xiangguo Zhang, Ruihao Gong, Shanghang Zhang, Qi~Zhang, Fengwei Yu, and Xianglong Liu. 2022.
\newblock Outlier suppression: Pushing the limit of low-bit transformer language models.
\newblock \emph{Advances in Neural Information Processing Systems}, 35:17402--17414.

\bibitem[{Wu et~al.(2023)Wu, He, Liu, Sun, Liu, Han, and Tang}]{wu2023brief}
Tianyu Wu, Shizhu He, Jingping Liu, Siqi Sun, Kang Liu, Qing-Long Han, and Yang Tang. 2023.
\newblock A brief overview of chatgpt: The history, status quo and potential future development.
\newblock \emph{IEEE/CAA Journal of Automatica Sinica}, 10(5):1122--1136.

\bibitem[{Xiao et~al.(2023)Xiao, Lin, Seznec, Wu, Demouth, and Han}]{xiao2023smoothquant}
Guangxuan Xiao, Ji~Lin, Mickael Seznec, Hao Wu, Julien Demouth, and Song Han. 2023.
\newblock Smoothquant: Accurate and efficient post-training quantization for large language models.
\newblock In \emph{International Conference on Machine Learning}, pages 38087--38099. PMLR.

\bibitem[{Xiao et~al.(2025)Xiao, Yang, Yang, Xu, Li, Su, Liu, Yang, and Wong}]{xiao2025ptqtp}
He~Xiao, Runming Yang, Qingyao Yang, Wendong Xu, Zhen Li, Yupeng Su, Zhengwu Liu, Hongxia Yang, and Ngai Wong. 2025.
\newblock Ptqtp: Post-training quantization to trit-planes for large language models.
\newblock \emph{arXiv preprint arXiv:2509.16989}.

\bibitem[{Yang et~al.(2025)Yang, Lin, Zhao, Wu, Zhu, Xie, Sun, Wang, and Gu}]{yang2025lrq}
Lianwei Yang, Haokun Lin, Tianchen Zhao, Yichen Wu, Hongyu Zhu, Ruiqi Xie, Zhenan Sun, Yu~Wang, and Qingyi Gu. 2025.
\newblock Lrq-dit: Log-rotation post-training quantization of diffusion transformers for image and video generation.
\newblock \emph{arXiv preprint arXiv:2508.03485}.

\bibitem[{Yao et~al.(2024)Yao, Duan, Xu, Cai, Sun, and Zhang}]{yao2024survey}
Yifan Yao, Jinhao Duan, Kaidi Xu, Yuanfang Cai, Zhibo Sun, and Yue Zhang. 2024.
\newblock A survey on large language model (llm) security and privacy: The good, the bad, and the ugly.
\newblock \emph{High-Confidence Computing}, 4(2):100211.

\bibitem[{Zellers et~al.(2019)Zellers, Holtzman, Bisk, Farhadi, and Choi}]{zellers2019hellaswag}
Rowan Zellers, Ari Holtzman, Yonatan Bisk, Ali Farhadi, and Yejin Choi. 2019.
\newblock Hellaswag: Can a machine really finish your sentence?
\newblock \emph{arXiv preprint arXiv:1905.07830}.

\bibitem[{Zhang et~al.(2022)Zhang, Roller, Goyal, Artetxe, Chen, Chen, Dewan, Diab, Li, Lin et~al.}]{zhang2022opt}
Susan Zhang, Stephen Roller, Naman Goyal, Mikel Artetxe, Moya Chen, Shuohui Chen, Christopher Dewan, Mona Diab, Xian Li, Xi~Victoria Lin, and 1 others. 2022.
\newblock Opt: Open pre-trained transformer language models.
\newblock \emph{arXiv preprint arXiv:2205.01068}.

\bibitem[{Zhang et~al.(2023)Zhang, Zhao, Lin, Sun, Yao, Han, Tanner, Liu, and Ji}]{zhang2023dynamic}
Yuxin Zhang, Lirui Zhao, Mingbao Lin, Yunyun Sun, Yiwu Yao, Xingjia Han, Jared Tanner, Shiwei Liu, and Rongrong Ji. 2023.
\newblock Dynamic sparse no training: Training-free fine-tuning for sparse llms.
\newblock \emph{arXiv preprint arXiv:2310.08915}.

\bibitem[{Zhou et~al.(2021)Zhou, Ma, Zhu, Liu, Zhang, Yuan, Sun, and Li}]{zhou2021learning}
Aojun Zhou, Yukun Ma, Junnan Zhu, Jianbo Liu, Zhijie Zhang, Kun Yuan, Wenxiu Sun, and Hongsheng Li. 2021.
\newblock Learning n: m fine-grained structured sparse neural networks from scratch.
\newblock \emph{arXiv preprint arXiv:2102.04010}.

\bibitem[{Zhu et~al.(2016)Zhu, Han, Mao, and Dally}]{zhu2016trained}
Chenzhuo Zhu, Song Han, Huizi Mao, and William~J Dally. 2016.
\newblock Trained ternary quantization.
\newblock \emph{arXiv preprint arXiv:1612.01064}.

\end{thebibliography}

\newpage
\appendix
\begin{figure*}[htbp]
    \centering    \includegraphics[width=0.9\linewidth]{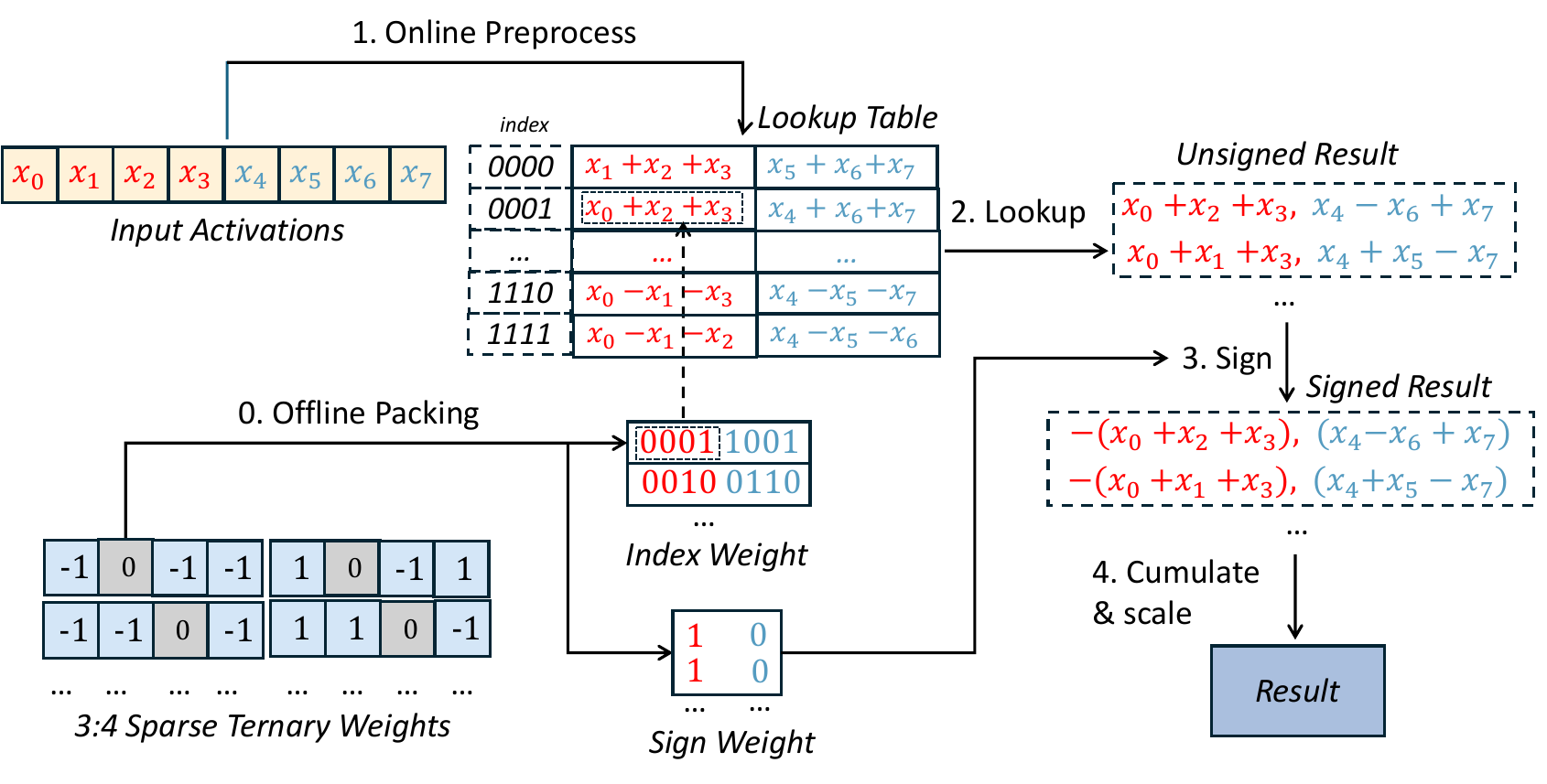}
    \vspace{-1em}
    \caption{The LUT-Based Inference engine for \textbf{Sherry}. The inference engine constructs a dynamic lookup table from input activations; weight indices are then used to fetch pre-computed values.}
    \label{fig: sys}
    \vspace{-1.5em}
\end{figure*}

\begin{center}
    \Huge\textbf{Appendix}
\end{center}

\section{Inference Engine Design}

\method seamlessly integrates with the BitNet.cpp~\cite{wang2025bitnetcpp} framework to enable efficient, multiplication-free inference. As illustrated in Fig.~\ref{fig: sys}, our engine operates in two distinct phases: an offline packing stage and an efficient online inference stage.

\paragraph{Offline Packing Phase}
During the offline phase, the 3:4 sparse ternary weights are compressed into a hardware-aligned format consisting of \textit{index} and \textit{sign} components. To maximize SIMD efficiency and memory density, every four-element block is packed into a 5-bit metadata structure: a 4-bit index represents the magnitude/sparsity pattern within the block, and a 1-bit value represents the shared or dominant sign. This packing strategy ensures that the weights are aligned with standard word boundaries, significantly reducing the bit-shuffling overhead compared to non-power-of-two schemes.

\paragraph{Online Inference Phase}
At inference time, the engine utilizes a high-performance lookup table (LUT) paradigm. The input activations are preprocessed into local LUTs for each weight segment. For each segment, the corresponding weight index serves as a direct pointer to retrieve pre-computed results from the LUT, entirely replacing floating-point multiplications with memory-efficient lookups. 

Following retrieval, the sign weight is applied to determine the final polarity of the segment sum. Subsequent accumulation across segments is performed via integer addition, concluding with the addition of the channel-wise scaling factor $\alpha$. This architecture results in a highly optimized inference path that realizes the theoretical efficiency of 1.25-bit quantization while requiring only minimal modifications to existing low-bit inference kernels.

\section{Related Work}
\label{sec: related}
\subsection{Quantization for LLMs}
Quantization has emerged as a cornerstone technique for enhancing the efficiency of Large Language Models (LLMs) by reducing the bit-precision of weights and activations~\citep{dettmers20218, dettmers2022llm, lin2023awq, frantar2022gptq, lin2025quantization, yang2025lrq, li2023loftq}. While effective, popular weight-only quantization methods~\citep{lin2023awq, frantar2022gptq} typically necessitate mixed-precision matrix multiplication, where weights and activations reside in disparate data formats. This heterogeneity requires specialized hardware support to maintain throughput, posing a significant hurdle for deployment on diverse edge and mobile platforms where hardware specialized for non-standard bit-widths is often unavailable.

To address this, weight-activation quantization strategies~\citep{dettmers2022llm, xiao2023smoothquant, huang2025quaff, wei2022outlier} seek a unified low-precision format. However, these methods often encounter the "outlier problem"~\citep{xiao2023smoothquant, dettmers2022llm}, where extreme activation values lead to high quantization errors. Consequently, it remains challenging for activations to reach the same ultra-low precision as weights without severe performance degradation, leaving a gap in achieving truly hardware-agnostic, high-efficiency deployment on the edge.

\subsection{Ternary Quantization}
Ternary quantization~\cite{li2016ternary, zhu2016trained}, often referred to as 1.58-bit quantization, offers a paradigm shift by constraining weights to the set $\{-1, 0, +1\}$. Beyond substantial memory reduction, this approach fundamentally simplifies the core matrix multiplication operation into addition, effectively eliminating the need for power-intensive multipliers~\cite{ma2025bitnet,wang2023bitnet}. This intrinsic hardware-friendliness is particularly vital for resource-constrained mobile environments.

Early research in this domain primarily focused on optimizing the quantization function through thresholding and scaling. Ternary Weight Networks (TWN)~\citep{li2016ternary} minimized reconstruction distortion by assuming a Gaussian weight distribution, while Trained Ternary Quantization (TTQ)~\citep{zhu2016trained} introduced learnable scaling factors to adapt the ternary representation during training. Subsequent work by \citet{leng2018extremely} leveraged the Alternating Direction Method of Multipliers (ADMM) to iteratively optimize these parameters.

The advent of Large Language Models (LLMs)~\cite{wu2023brief, floridi2020gpt, zhang2022opt} has revitalized interest in ternary schemes, as the generative capacity of LLMs is notoriously sensitive to precision loss. This has led to two distinct research trajectories. The first utilizes Post-Training Quantization (PTQ)~\cite{lin2023awq, frantar2022gptq, xiao2025ptqtp} to minimize the computational cost of quantization; however, PTQ often suffers from non-negligible performance drops at ultra-low bit-widths. Consequently, Quantization-Aware Training (QAT)~\cite{chen2025efficientqat} has become the preferred choice for robust performance recovery. Within the QAT landscape, strategies range from the straightforward AbsMean scaling utilized by the BitNet family~\cite{ma2025bitnet, wang2023bitnet, wang2025bitnetcpp} to more sophisticated adaptations of Learned Step Size Quantization (LSQ)~\cite{esser2019learned} for the ternary regime~\cite{chen2024ternaryllm, liu2025paretoq}. 

Regarding the inference engine, classical frameworks like \texttt{llama.cpp} still rely on multiplication-based kernels that require weight unpacking and floating-point operations. To overcome this, T-Mac~\cite{wei2025t} and TENET~\cite{huang2025tenet} proposed a lookup table (LUT)-based engine to eliminate multiplications, though it requires packing ternary weights into 2-bit containers. BitNet.cpp~\cite{wang2025bitnetcpp} attempted to reduce the footprint further via a 1.67-bit packing strategy (3 weights in 5 bits). However, this 3-way packing is inherently SIMD-unfriendly, introducing significant bit-shuffling overhead that often results in slower performance than 2-bit packing.

Consequently, existing ternary models are forced to trade off between bit width and inference speed. \method addresses this dilemma by introducing hardware-aligned 3:4 structured sparsity, achieving a 1.25-bit SIMD-friendly packing.
\begin{figure*}[t!]
    \centering    \includegraphics[width=0.9\linewidth]{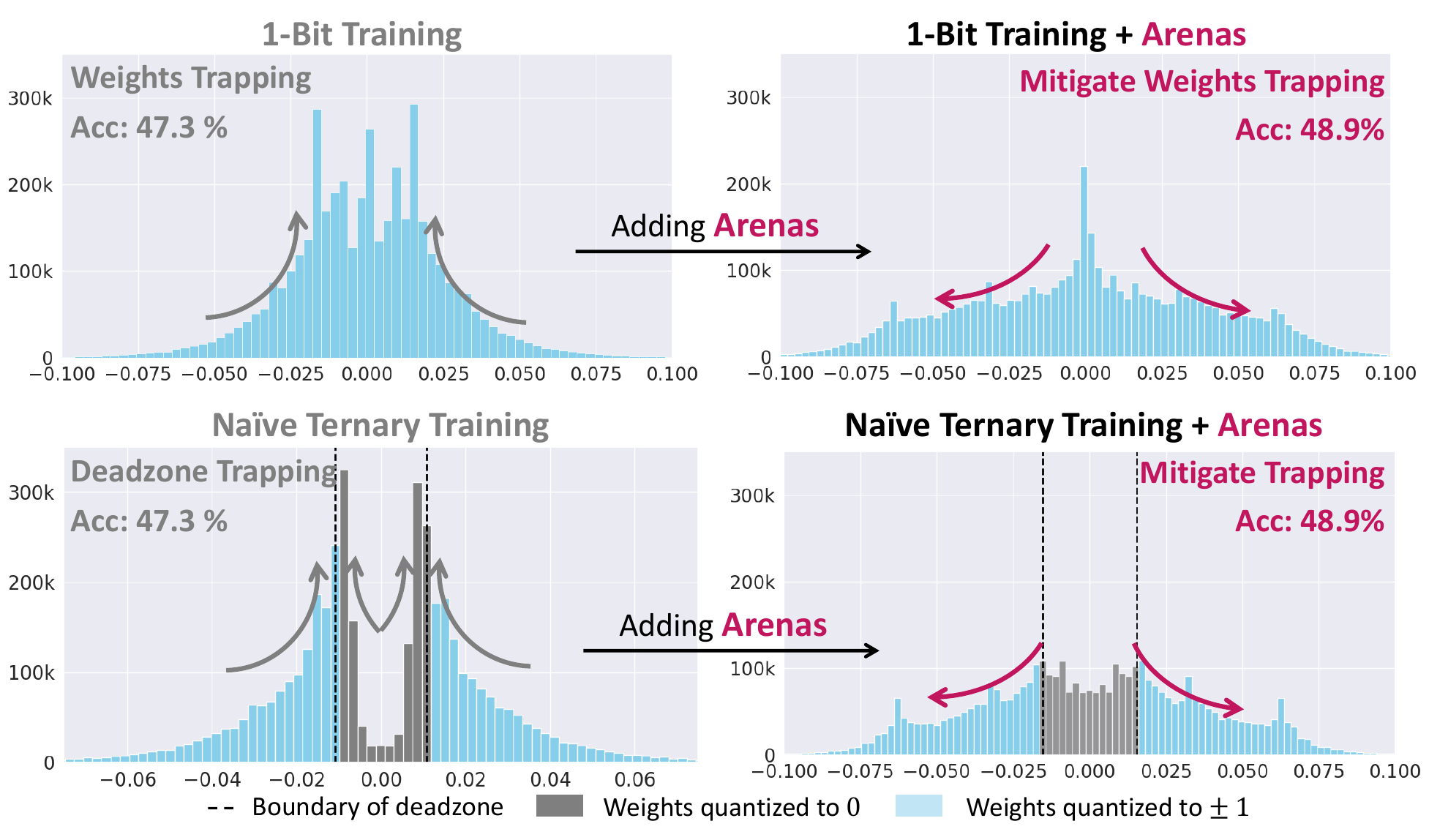}
    \vspace{-1em}
    \caption{Weight distributions across different quantization regimes. The inclusion of Arenas effectively mitigates the trapping phenomenon in both binary and naive ternary quantization.}
    \label{fig: more_arenas}
    \vspace{-1.5em}
\end{figure*}
\subsection{N:M sparsity}
N:M structured sparsity~\cite{lin2023efficient, zhou2021learning} has emerged as a critical middle ground between the flexibility of unstructured pruning and the hardware efficiency of block-wise pruning. By enforcing that exactly $N$ out of every $M$ contiguous weights are non-zero, this pattern provides predictable memory access patterns highly amenable to hardware acceleration. Notably, NVIDIA’s Ampere and subsequent architectures~\cite{lin2023efficient} introduced native Tensor Core support for 2:4 sparsity, doubling throughput with minimal accuracy loss in high-precision models.

Existing research~\cite{sun2021dominosearch, fu2023effectiveness, zhang2023dynamic} on N:M sparsity has primarily focused on the selection of optimal masks and performance recovery via specialized training regimes. However, most efforts are not coordinated with ultra-low bit quantization, as they are largely designed for Sparse Tensor Cores on GPUs, which currently prioritize 16-bit or 32-bit floating-point arithmetic.

In the context of ultra-low bit-widths, the intersection of N:M sparsity and ternary quantization remains largely unexplored. Furthermore, as the bit-width decreases, the rigid N:M constraint exacerbates the \textbf{weight trapping} phenomenon. \method extends the N:M paradigm by introducing a hardware-aligned 3:4 pattern coupled with the Arenas module. This synergy is specifically designed to bridge the accuracy gap between sparse and dense ternary models with zero additional inference overhead. \textit{To our knowledge, Sherry is the first to provide a hardware-efficient solution for applying N:M sparsity on ultra-low bit-width quantization.}

\section{Optimality of the 3:4 Sparse Format}
\label{sec: ideal}

We argue that the 3:4 structured sparsity pattern represents the optimal solution among all $N:M$ ($N < M$) formats for LUT-based ternary inference engines. To formalize this, we define the search space as packing $M$ ternary weights into $B$ bits, where $B/M$ determines the effective bit-width. In a hardware-aligned implementation, these $B$ bits are partitioned into 1 sign bit and $B-1$ index bits (utilizing the mirror-symmetry of ternary states).

\subsection{Design Constraints}
The selection of $N$ and $M$ is governed by three critical hardware and performance constraints:

\begin{enumerate}
    \item[(1)] \textbf{SIMD Alignment:} The block size $M$ must be a power of two ($M \in \{2^k\}$) to ensure predictable memory access and alignment with standard SIMD register widths.
    \item[(2)] \textbf{LUT Capacity:} Standard SIMD instructions (e.g., x86 AVX2 vpshufb) utilize a 128-bit (16-byte) register as a lookup table. This constrains the index to 4 bits ($2^4 = 16$ entries), implying $B-1 \le 4$.
    \item[(3)] \textbf{Sparsity Threshold:} To preserve representational capacity and avoid severe accuracy degradation~\cite{zhu2016trained}, the density ratio must satisfy $\frac{N}{M} \ge 0.5$ (i.e., sparsity $\le 50\%$).
\end{enumerate}

\subsection{Proof of Optimality}
By applying hardware Constraints (1) and (2), the search space for $M$ is limited to $\{2, 4, 8\}$. For a target bit-width of approximately 1.25 bits, we evaluate the following candidate configurations:

\begin{itemize}
    \item \textbf{M=2 (1:2):} Requires $B=4$ bits ($B/M = 2.0$), failing the efficiency requirement.
    \item \textbf{M=8 (5:8, 6:8, 7:8):} These require $B \ge 10$ bits to maintain efficiency, leading to $B-1 \ge 9$. This violates Condition (2), the 4-bit LUT capacity of a single SIMD instruction, requiring multi-cycle split-table lookups.
    \item \textbf{M=4 (2:4, 3:4):} These are the only candidates satisfying both hardware constraints.
\end{itemize}

To choose between 2:4 and 3:4, we consider the information density of the index space. A 3:4 ternary block with one shared sign bit has a total of $C_4^3 \cdot 2^{3-1} = 16$ unique patterns. This perfectly saturates the $2^4$ entries in the LUT. In contrast, a 2:4 scheme only utilizes $C_4^2 \cdot 2^{2-1} = 12$ states, resulting in bit-waste. 

Furthermore, 2:4 sparsity resides exactly on the 50\% threshold where performance begins to destabilize. The 3:4 format provides 75\% density (25\% sparsity), staying well within the safe margin for model expressive capacity while maximizing bit-utilization. Thus, the 3:4 format is the optimal solution for high-performance 1.25-bit SIMD inference.

\section{Optimality of the Sparse-Absmean}
\label{sec: proof}
Let $W_{b:b+3, j} = (W_{b,j}, W_{b+1,j},W_{b+2,j},W_{b+3,j})$ denote a contiguous vector. The objective is formulated as:
\begin{equation}
\begin{aligned}
\min_{T, \alpha} & \sum_{j=1}^{d_{out}} \|W_{:, j} - T_{:, j} \alpha_j \|_2^2 \\
\text{s.t. } & T_{i,j} \in \{-1, 0, +1\}, \\
& \forall b\in \{1, 5,\dots, d_{in}-4 \}: \|T_{b:b+3, j}\|_0 = 3.
\end{aligned}
\end{equation}

where $\|\cdot\|_0 = 3$ enforces the 3:4 structured sparsity by ensuring exactly three non-zero ternary values per block. 

To determine the optimal indices and signs for $T$, we expand the per-block objective for a block $b$:
\begin{equation}
    \min_{T, \|T\|_0=3} \sum_{i=b}^{b+3} (W_{i,j}^2 - 2 \alpha_j W_{i,j} T_{i,j} + \alpha_j^2 T_{i,j}^2).
\end{equation}
Since $\sum T_{i,j}^2 = 3$ is constant for any valid 3:4 pattern, minimizing the error is equivalent to maximizing the correlation term:
\begin{equation}
    \max_{T, \|T\|_0=3} \sum_{i=b}^{b+3} W_{i,j} T_{i,j}.
\end{equation}
To maximize this sum, we must choose $T_{i,j} = \text{sign}(W_{i,j})$ for the three elements with the largest absolute magnitudes and set $T_{i,j} = 0$ for the element with the smallest absolute magnitude in the block.

\paragraph{Sign Optimality:} For any element $i$ where $T_{i,j} \neq 0$, the quadratic term $(W_{i,j} - T_{i,j}\alpha)^2$ is minimized when $T_{i,j}$ has the same sign as $W_{i,j}$. If $\text{sign}(T_{i,j}) \neq \text{sign}(W_{i,j})$, flipping the sign of $T_{i,j}$ would increase the correlation $W_{i,j}T_{i,j}$ and strictly decrease the objective for any $\alpha > 0$.

\paragraph{Index Optimality:} Let $\{|W|_{(1)}, |W|_{(2)}, |W|_{(3)},$ $ |W|_{(4)}\}$ be the sorted absolute values of weights in a 4-element block such that $|W|_{(1)} \geq |W|_{(2)} \geq |W|_{(3)} \geq |W|_{(4)}$. The correlation sum for the block is $\sum_{k \in \mathcal{I}} |W|_{(k)}$, where $\mathcal{I}$ is the set of indices of the 3 chosen elements. To maximize this sum, we must choose the indices corresponding to the three largest magnitudes: $(1), (2),$ and $(3)$.

By combining the sign and index selection, for each block $i\in[b,b+3]$, the optimal ternary element $T_{i,j}$ is given by:
\begin{equation}
T^*_{i,j} = \begin{cases} 
0, & \text{if } i = \arg\min |W_{i,j}|; \\ 
\text{sign}(W_{i,j}), & \text{otherwise}.
\end{cases}
\end{equation}
Substituting $T_{i,j}^*$ back into the expression for $\alpha_j^*$ gives the global minimum:
\begin{equation}
\alpha^*_j = \frac{4}{3 d_{in}} \sum_{i \in \mathcal{S}_j} |W_{i,j}|,
\end{equation}
where $\mathcal{S}_j = \{i \mid T_{i,j} \neq 0\}$ denotes the set of active (non-zero) indices in the $j$-th column. 

\section{Ternary Quantization Baselines}
\label{sec: baseline}
Recall the general form of ternary quantization: taking per-channel quantization as an example, for a full-precision weight matrix $W \in \mathbb{R}^{d_{in} \times d_{out}}$, the general ternary quantization function $Q(\cdot)$ is defined as:
\begin{equation}
    Q(W) = T\alpha, \quad T_{i,j} = \begin{cases} 
    +1, & \text{if } W_{i,j} > \Delta_j; \\ 
    0, & \text{if } |W_{i,j}| \le \Delta_j; \\ 
    -1, & \text{if } W_{i,j} < -\Delta_j, 
    \end{cases} 
\end{equation}
where $T \in \{-1, 0, +1\}^{d_{in} \times d_{out}}$ is the ternary weight matrix, $\alpha \in \mathbb{R}^{d_{out}}$ represents the scaling factors, and $\Delta \in \mathbb{R}^{d_{out}}$ denotes the quantization thresholds.

\begin{figure*}[t]
    \centering    \includegraphics[width=0.9\linewidth]{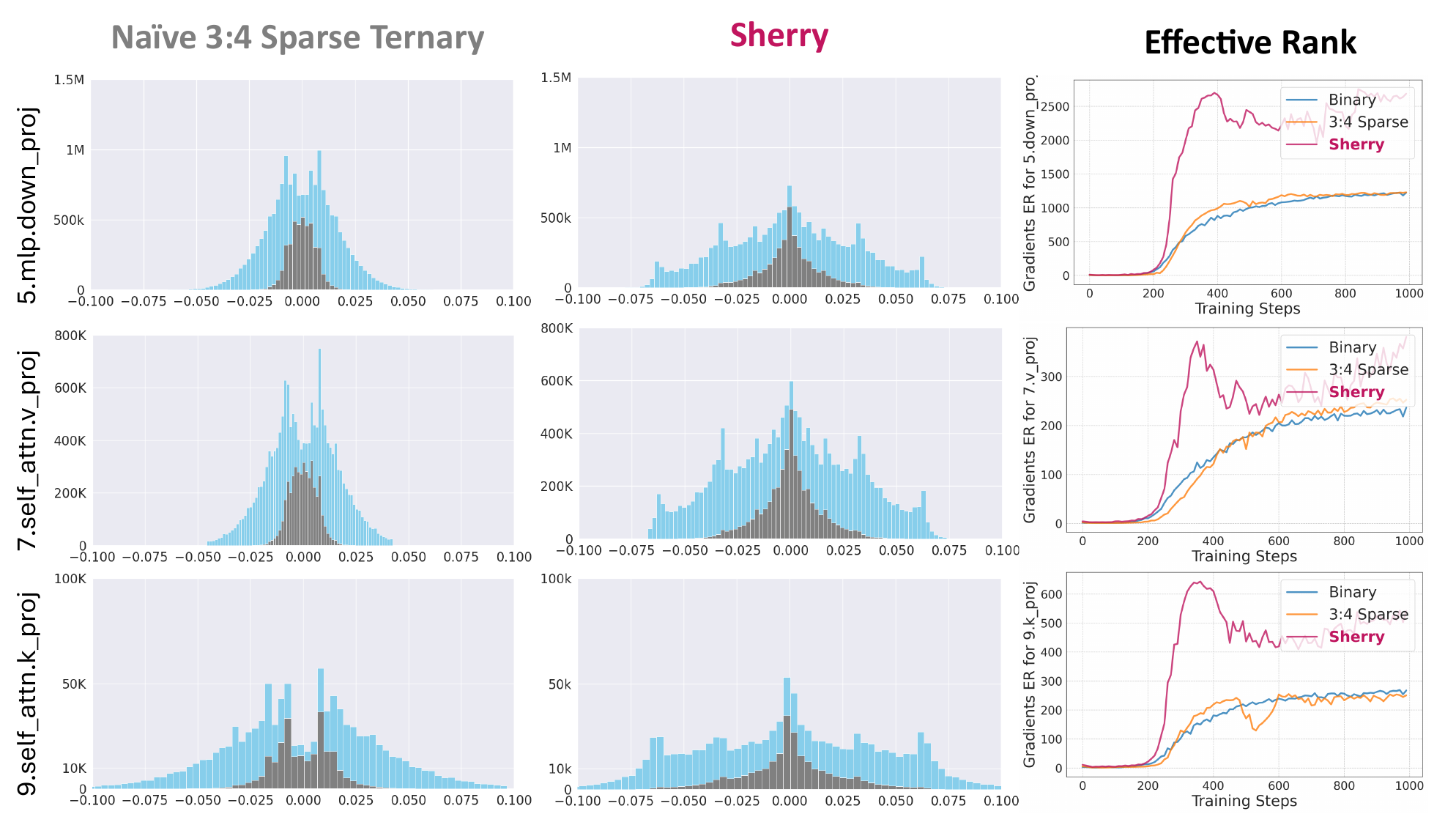}
    \caption{Weight distributions and the Effective Ranks (ER) of their gradients across different layers.}
    \label{fig: more_distribution}
    \vspace{-1.5em}
\end{figure*}

For contemporary large-scale ternary models~\citep{ma2025bitnet, team2025minicpm4}, the AbsMean method has emerged as the de facto standard due to its superior training stability. In AbsMean, the scaling factor $\alpha_j$ and threshold $\Delta_j$ for the $j$-th channel are calculated as:
\begin{equation}
    \alpha_j = \frac{1}{d_{in}}\sum_{i=1}^{d_{in}} |W_{i,j}|, \quad \Delta_j = \frac{\alpha_j}{2}.
    \label{eq:absmean_baseline}
\end{equation}

Besides Absmean, previous optimization strategies can be broadly categorized into: (1) reducing quantization error and (2) enhancing model expressive capacity.

\subsection{Reducing Quantization Error}
A primary line of work focuses on optimizing $\Delta$ and $\alpha$ to minimize the reconstruction error 
\begin{equation}
   \min_{\Delta, \alpha} \|W - T\alpha\|^2_2. 
\end{equation}
This is exemplified by estimation-based methods like Ternary Weight Networks (TWN)~\citep{li2016ternary}, which minimize distortion by assuming a Gaussian weight distribution to approximate 
\begin{equation}
    \Delta^*_j \approx 0.7 \cdot \mathbb{E}[|W_{:,j}|].
\end{equation}
For a fixed $\Delta$, the optimal $\alpha$ is: 
\begin{equation}
\alpha^*_j = \frac{1}{|\mathcal{S}_j|} \sum_{i \in \mathcal{S}_j} |W_{i,j}|,
\end{equation}
where $\mathcal{S}_j$ is the set of weights $W_{i,j}$ exceeding $\Delta_j$. However, the Gaussian assumption often fails in LLMs, leading to biased estimates and performance degradation. While subsequent methods like LSQ~\citep{esser2019learned} and DLT~\citep{chen2024ternaryllm} introduce trainable scaling factors, they are easier to being trapped in sub-optima, exhibiting slower convergence and higher final loss compared to AbsMean~\cite{huang2025tequila}.

\subsection{Enhancing Expressive Capacity}
Alternative approaches attempt to recover capacity by incorporating bias terms. DLT~\cite{chen2024ternaryllm} introduces a learnable bias during dequantization:
\begin{equation}
Y = X(T\alpha + B) = XT\alpha + Xb.
\end{equation}
However, this necessitates a dense full-precision multiplication ($Xb$), destroying the computational efficiency of ternary quantization. Similarly, SEQ~\cite{liu2025paretoq} reassigns the zero-point to a non-zero value $\alpha_j b_j$:
\begin{equation}
    T_{i,j} = \begin{cases} 
    +1, & \text{if } W_{i,j} > \Delta_j; \\ 
    \alpha_j b_j, & \text{if } |W_{i,j}| \le \Delta_j; \\ 
    -1, & \text{if } W_{i,j} < -\Delta_j, 
    \end{cases} 
    \label{eq:tquant}
\end{equation}
This also voids hardware efficiency, as the resulting operations are no longer multiplication-free. In contrast, \method maintains strict ternary efficiency during inference by utilizing an annealing residual that vanishes post-training.

\section{Effective Rank for Gradient Analysis}
\label{sec: er}

To quantitatively assess the diversity and information density of the gradient updates, we utilize the \textbf{Effective Rank} ($ER$) metric~\cite{roy2007effective}. While the standard algebraic rank provides a binary measure of linear independence, it is highly sensitive to infinitesimal noise and fails to capture the numerical stability of the gradient matrix. In contrast, Effective Rank provides a continuous measure of the "learning dimensionality" of a layer by evaluating the entropy of its singular value distribution.

Given a weight gradient matrix $G \in \mathbb{R}^{d_{in} \times d_{out}}$, let $\sigma = (\sigma_1, \sigma_2, \dots, \sigma_k)$ denote its singular values obtained via Singular Value Decomposition (SVD), sorted in descending order. We first normalize these values to form a probability-like distribution $p$:
\begin{equation}
    p_i = \frac{\sigma_i}{\sum_{j=1}^{k} \sigma_j}, \quad i = 1, \dots, k.
\end{equation}
The Effective Rank is then defined as the exponential of the Shannon entropy of this distribution:
\begin{equation}
    ER(G) = \exp \left( -\sum_{i=1}^{k} p_i \ln p_i \right).
\end{equation}

The value of $ER(G)$ ranges from $1$ to $k$. An $ER$ close to $1$ indicates that the gradients are highly \textit{homogenized}, with the update signal collapsing into a single dominant direction. Conversely, an $ER$ close to $k$ suggests a high-rank, diverse update where many independent features are being learned simultaneously.

In the context of N:M sparsity, we use $ER$ to diagnose \textbf{Gradient Homogenization}. As shown in Figure~\ref{fig: er} and~\ref{fig: more_distribution}, naive 3:4 training regimes often exhibit a relatively small $ER$ as weights become "trapped" in discrete slots, leading to redundant updates. By maintaining a higher $ER$, \method ensures that the sparse ternary model retains the representational richness of its dense counterpart.

\section{More Experimental Details and Results}
\label{sec: setup}

\subsection{Evaluation Benchmarks}
\label{sec: metrics}

\paragraph{PIQA:} The Physical Interaction Question Answering (PIQA) benchmark~\cite{bisk2020piqa} focuses on physical commonsense reasoning. It tests a model's understanding of everyday physical laws by presenting scenarios (e.g., "How do you stabilize a wobbly table?") and requiring the selection of the correct solution from two options. Success on PIQA indicates a foundational grasp of physical object interactions.

\paragraph{ARC-Easy and ARC-Challenge:} The AI2 Reasoning Challenge (ARC) dataset~\cite{clark2018think} is bifurcated to assess scientific knowledge. The ARC-Easy set contains grade-school science questions often answerable via fact retrieval. Conversely, ARC-Challenge consists of questions curated to be difficult for standard retrieval algorithms, requiring complex multi-step reasoning and a deeper conceptual understanding.

\paragraph{HellaSwag:} HellaSwag~\cite{zellers2019hellaswag} evaluates contextual commonsense inference. Models are given a situational premise and must select the most plausible continuation from four options. Crucially, distractors are adversarially generated to be deceptively plausible, ensuring that success requires a nuanced understanding of event dynamics rather than simple word association.

\paragraph{WinoGrande:} WinoGrande~\cite{sakaguchi2021winogrande} is a large-scale dataset for assessing commonsense reasoning through pronoun resolution. Derived from the Winograd Schema Challenge, it presents sentences with ambiguous pronouns  and requires the model to determine the correct referent. WinoGrande employs adversarial filtering to reduce statistical biases, forcing models to rely on genuine commonsense understanding.

\subsection{Annealing Gate Schedule $\lambda_t$}

The hyperparameter $\lambda_t$ plays a crucial role in the annealing gate. Proper scheduling of $\lambda_t$ from an initial value of 1 to a final value of 0 throughout the training process can significantly improve model performance and convergence stability. This section describes three decay strategies in Fig.~\ref{fig: schedule} for $\lambda_t$ scheduling, where $p_t \in [0,1]$ represents the training progress (0: training start, 1: training completion).

The linear decay strategy provides a constant reduction rate of $\lambda_t$ throughout the training process:
\begin{equation}
\label{eq:linear_decay}
\lambda_t = 1 - p_t .
\end{equation}

The cosine decay strategy follows a convex curve, with slower decay at the beginning and end of training, and faster decay in the middle stage:
\begin{equation}
\label{eq:cosine_decay}
\lambda_t = \frac{1}{2}\left[1 + \cos(\pi p_t)\right] .
\end{equation}

The exponential decay strategy follows a concave curve, with rapid decay at the beginning that gradually slows down:
\begin{equation}
\label{eq:exponential_decay}
\lambda_t = \exp(-5 p_t)
\end{equation}

\section{Broader Impact}
The development of \method carries significant implications for the democratization and sustainability of Large Language Models (LLMs). By achieving high-performance 1.25-bit inference through hardware-aligned 3:4 structured sparsity, our work opens up significant societal and technological benefits in resource-constrained scenarios~\cite{li2025prima, huang2023distributed, huang2024fedmef}, such as edge/mobile computing~\cite{forman1994challenges, imielinski1996mobile, chen2024distributed} and cross-device federated learning~\cite{mcmahan2017communication, huang2025fedrts}. This empowers researchers and developers in resource-constrained environments and developing regions to access and innovate with large-scale models without prohibitive hardware costs.

\end{document}